\documentclass{article}
\usepackage{fullpage}
\usepackage{ifthen}
\usepackage[round, sort]{natbib}
\usepackage{fancyhdr}
\usepackage[format=plain,
            labelfont={it,it},
            textfont=it]{caption}
\usepackage[stable, bottom, flushmargin]{footmisc}
\usepackage{graphicx}
\usepackage[pagebackref, colorlinks=true, linkcolor=blue, citecolor=cyan]{hyperref}
\usepackage{amssymb}

\usepackage{microtype}
\usepackage{graphicx}
\usepackage{booktabs} 

\usepackage{amsmath}
\usepackage{mathtools}
\usepackage{amsthm}

\usepackage{blkarray}
\usepackage{bbm}
\usepackage{mathtools}
\usepackage{makecell}
\usepackage{bm}
\usepackage{subfig}
\usepackage{graphics}
\usepackage{listings}
\usepackage{multirow}
\usepackage{xcolor}
\usepackage{comment}
\usepackage{amsfonts}

\usepackage{xspace}
\usepackage{thmtools} 
\usepackage{thm-restate}
\usepackage{pifont}
\usepackage{booktabs}
\usepackage{balance}
\usepackage[capitalize,noabbrev]{cleveref}

\declaretheorem[name=Proposition]{proposition}
\declaretheorem[name=Theorem, sharenumber=proposition]{theorem}

\declaretheorem[name=Definition, sharenumber=proposition]{definition}
\declaretheorem[name=Corollary, sharenumber=proposition]{corollary}

\declaretheorem[name=Lemma, sharenumber=proposition]{lemma}

\newcommand{\X}{\mathcal{X}}

\renewcommand{\P}[1]{\mathcal{P}^{(#1)}}

\newcommand{\D}{\mathcal{D}}

\newcommand{\M}{M}

\newcommand{\R}{\mathbbm{R}}

\newcommand{\1}{\mathbbm{1}}

\newcommand{\lab}[1]{\lambda_{#1}}
\newcommand{\atrank}[2]{\mathcal{R}_{#1,#2}}
\newcommand{\nlab}[2][]{N_{#1}^{#2}}
\newcommand{\dTV}[2]{d_{\text{TV}}(#1,#2)}

\providecommand{\half}{\ensuremath{\frac{1}{2}}\xspace}

\providecommand{\SET}[1]{\ensuremath{\left\{ #1 \right\}}\xspace}

\providecommand{\Set}[2]{\ensuremath{\SET{#1 \mid #2}}\xspace}

\providecommand{\Kth}[1]{\ensuremath{{#1}^{\rm th}}}

\providecommand{\PROB}{\ensuremath{\mathbbm{P}}\xspace}
\providecommand{\Prob}[2][]{\ensuremath{%
\ifthenelse{\equal{#1}{}}{\PROB[#2]}{\PROB_{#1}[#2]}}\xspace}
\providecommand{\ProbC}[3][]{\Prob[#1]{#2\;|\;#3}}
\providecommand{\Expect}[2][]{\ensuremath{%
\ifthenelse{\equal{#1}{}}{\mathbb{E}}{\mathbb{E}_{#1}}%
\left[#2\right]}\xspace}

\newcommand{\LPlabel}{}

\newcommand{\one}[1]{\1(#1)}

\renewcommand{\R}{\mathbb{R}}
\renewcommand{\M}{\mathcal{M}}

\newcommand{\dsmatrix}{\M^{n\times n}_{\text{DS}}}

\DeclareMathOperator{\Ima}{Im}
\newcommand{\RUP}[1][]{r^{#1}_{\text{UA}}}
\newcommand{\rup}[2][]{\RUP[#1](#2)}

\renewcommand{\P}{\mathcal{P}}

\newcommand{\ropt}{r^{\tau}_{\textrm{opt}}}
\newcommand{\ROPT}[1][]{r^{\tau}_\textrm{opt}}
\newcommand{\ropteq}[2][]{\ROPT[#1](#2)}
\newcommand{\rmin}{r^{\tau}_{\textrm{min}}}

\newcommand{\rua}{r_\textrm{UA}}
\newcommand{\runif}{r_\textrm{unif}}
\newcommand{\rmix}{r^\phi_{\textrm{mix}}}
\newcommand{\rpl}{r_\textrm{PL}}

\newcommand{\email}[1]{\textsf{#1}}
\newcommand\blfootnote[1]{%
  \begingroup
  \renewcommand\thefootnote{}\footnote{#1}%
  \addtocounter{footnote}{-1}%
  \endgroup
}

\title{Stability and Multigroup Fairness in Ranking with\\ Uncertain Predictions}

\author{Siddartha Devic \\ University of Southern California \\ \email{devic@usc.edu} \and 
Aleksandra Korolova\footnotemark[1] \\ Princeton University\\ \email{korolova@princeton.edu} \and
David Kempe\footnotemark[1] \\ University of Southern California \\ \email{david.m.kempe@gmail.com} \and
Vatsal Sharan\footnotemark[1] \\ University of Southern California \\ \email{vsharan@usc.edu}
}
\date{}

\begin{document}
\maketitle

\blfootnote{{*} Equal contribution; the order of these authors was randomized with \url{https://www.random.org/}.}

\begin{abstract}
Rankings are ubiquitous across many applications, from search engines to hiring committees.
In practice, many rankings are derived from the output of predictors.
However, when predictors trained for classification tasks have intrinsic uncertainty, it is not obvious how this uncertainty should be represented in the derived rankings. 
Our work considers \emph{ranking functions}: maps from individual predictions for a classification task to distributions over rankings.
We focus on two aspects of ranking functions: stability to perturbations in predictions and fairness towards both individuals and subgroups.
Not only is stability an important requirement for its own sake, but --- as we show --- it composes harmoniously with individual fairness in the sense of \citet{dwork2012fairness}.
While deterministic ranking functions cannot be stable aside from trivial scenarios, we show that the recently proposed \emph{uncertainty aware} (UA) ranking functions of \citet{singh2021fairness} are stable.
Our main result is that UA rankings also achieve multigroup fairness through successful composition with \emph{multiaccurate} or \emph{multicalibrated} predictors.
Our work demonstrates that UA rankings naturally interpolate between group and individual level fairness guarantees, while simultaneously satisfying stability guarantees important whenever machine-learned predictions are used.
\end{abstract}

\section{Introduction}
Rankings underpin many modern systems: companies rank job applications \citep{turboHireRank, geyik2019fairness}, ad marketplaces rank ads to serve to a user \citep{googleAd}, and social media platforms and feeds rank content \citep{facebookRank}.
Rankings are also used to partially automate decision making in settings with limited resources or attention span (such as job candidate interview selections or ad delivery).
Rankings are often derived from predictions generated by machine learning models designed and deployed on relevant classification tasks.
For example, a job advertisement platform may use a model which predicts an individual's \emph{relevance score} for each job they apply to; say, on a scale from 1 to 3 (corresponding to irrelevant, suitable, or extremely relevant); this score is then factored into the platform's ranking of job applicants shown to a company recruiter with a limited time budget. 

In practice, machine learning models often predict \emph{distributions} over classes instead of a single class.
This is because predictions correspond to a belief about what the future may possibly hold, but not a certainty about what the future \emph{will} look like.
A plethora of recent research in model calibration \citep{guo2017calibration, minderer2021revisiting, gupta2022toplevelcalibration}, conformal prediction \citep{jung2022practicaladversarial,jung2023batchmultivalid}, and uncertainty quantification \citep{angelopoulos2021gentle} has tackled the issue of ensuring that the uncertainty estimates output by a model are meaningful, rather than artifacts of any particular training regime.

With uncertainty inherent in predictions, we argue that it is essential to revisit the question of how to meaningfully convert predictions (made in the form of distributions over classes) into rankings.
Without any uncertainty --- for example, if one had access to an oracle for the future --- it would usually be clear what a ranking should look like: \emph{meritocracy} would suggest that one always places the more suitable candidates higher in the ranking.
However, when given only \emph{predictions} about suitability/merit with intrinsic uncertainty, the approach for generating a meaningful ranking is less clear.
After all, one must choose a ranking over candidates, e.g., in order to make an interview decision, \emph{before} witnessing the exact suitability of each candidate,
which is generally only observable after an individual works in the job for months or even years.
Since an uncertain prediction can be considered a \emph{prior} (and typically imperfect) belief on the qualifications or performance of any given individual, the fundamental task of designing a meaningful ranking algorithm \emph{utilizing} these predictions must be reexamined.

For a meaningful derivation of rankings from predictors, we consider the following two properties to be essential requirements:

\paragraph{Anonymity.} All individuals must be treated symmetrically a priori, i.e., if the predictions are permuted, then the ranking is permuted according to the same permutation.

\paragraph{Stability.} If the predictor's distribution over classes changes only slightly (in Total Variation distance), then the corresponding induced ranking should only change slightly.
\newline

The reason for requiring anonymity is self-evident: it rules out discrimination on the basis of the identity of individuals.
Stability is more nuanced: it articulates a desire to have rankings which are agnostic to small amounts of noise in the predictions for each individual.
In deployed applications, a small amount of variation injected by a seeded training/test data set split or a randomized training procedure can introduce noise at the level of individual predictions \citep{ganesh2023impact}.
Furthermore, there will also always be at least some additional noise due to incomplete data entries, mistaken inputs, etc.~\citep{nettleton2010study}.
Rankings should be generally agnostic to these sources of noise: if minute noise in predictions can induce large changes in the derived ranking, the ranking cannot be very meaningful or fair to begin with.
Stability can therefore be interpreted as a way to combat micro-\emph{arbitrariness} of rankings induced by learned predictors \citep{cooper2023my}.
For stability to be meaningful, we will need to shift our focus to \emph{distributions} over rankings: utilizing randomness to deal with uncertainty will be key in achieving stability.

Anonymity can be construed as a fairness notion, but it is a very minimal one.
Fairness in a stronger sense has been the focus of much recent work, both in the context of ranking (see, e.g., \citet{singh2018fairness, singh2019policy}) and in the context of classifiers/predictors (see, e.g., \citet{hardt2016equality, caton2020fairness, dwork2012fairness, awasthi2020equalized}).
As ML-based predictors are often used in order to ultimately produce rankings \citep{wang2012learning}, it is a natural desideratum that the ranking function \emph{preserve} fairness guarantees of the underlying predictor: this ensures that no additional unfairness is introduced in post-processing the classifier's output. 
As we will see, not all ranking functions satisfy such fairness \emph{composition} properties.

\subsection{Our Contributions}

\begin{figure*}[ht]
\begin{centering}
  \includegraphics[scale=0.24]{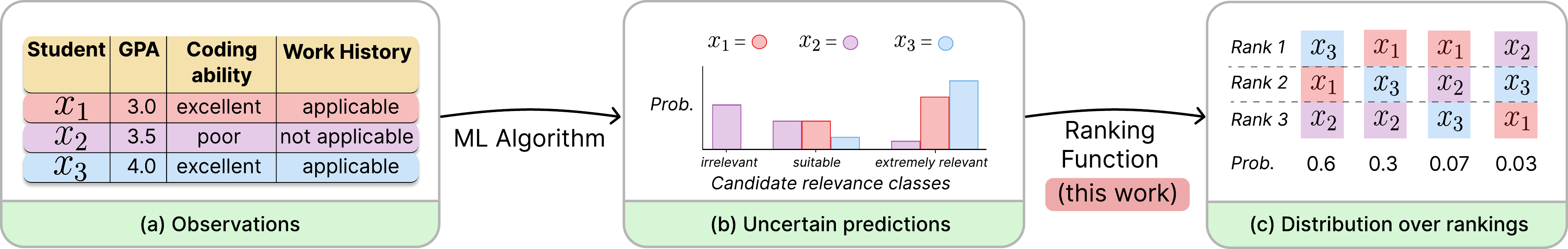}
  \caption{
  An overview of our setting, using students being ranked by an employer for potential interviews. Observations (a), given by the students' resum\'es and coding abilities, are fed into a machine learning algorithm which produces distributions (b) over the
  relevance classes \{irrelevant, suitable, extremely relevant\} for each candidate. Then, a \emph{ranking function} takes as input these uncertain predictions to produce a distribution (c) over rankings of the three candidates. 
  Although it may appear that $x_3$ is the most qualified or relevant, due to inherent uncertainty in observations, a ranking function may place $x_1$ or $x_2$ at rank one with non-zero probability (c).
  }
  \vspace{-1.2em}
  \label{fig:main_fig}
\end{centering}
\end{figure*}

We focus on scenarios in which individuals, scored by a predictor, must be presented to a decision maker in a linear order or \emph{ranking}.
We assume that the predictions take the form of \emph{distributions over classes}, modeling inherent uncertainty in the underlying ground truth or data.
In \cref{sec:preliminaries}, we define a \emph{ranking function} as a map from such probabilistic predictions to a distribution over rankings.
\cref{fig:main_fig} illustrates this setting with an example.

Our first (and very immediate --- see \cref{sec:stability-consequences}) result is that stability naturally composes with individual fairness \citep{dwork2012fairness}: if the predictor is individually fair and the ranking function is stable, then the composition satisfies a natural generalization of individual fairness to rankings. 
This result further confirms that stability is a desirable property for a ranking function.

In light of the desirability of stability, we next investigate which ranking functions are stable. Deterministic ranking functions are natural and popular; unfortunately, we show (\cref{sec:deterministic}) that the only \emph{stable} deterministic ranking functions are constant, i.e., trivial functions that output the same ranking regardless of the predictions. Further, deterministic ranking functions cannot be anonymous. Thus, one must choose between stability/anonymity and determinism, providing significant evidence in favor of randomization.
With randomization, stability and anonymity both become achievable: we show (\cref{sec:up-rankings}) that a natural adaptation of the Uncertainty Aware (UA) ranking functions of \citet{singh2021fairness, devic23Fairness} to the case of multiclass predictions of the classifier is indeed anonymous and stable.

We then investigate the fairness guarantees of UA ranking in more depth, and prove (\cref{sec:multicalibration-guarantees}) our main result: UA ranking naturally preserves multiaccuracy and multicalibration guarantees\footnote{
Multiaccuracy requires that the uncertainty estimates of a predictor are \emph{unbiased} over a set of subgroups (combating discrimination between groups); multicalibration guarantees that the estimates are also \emph{calibrated} over subgroups (combating discrimination between \emph{and within} groups).
These are arguably the most popular notions of fairness in settings with uncertain predictions where predictors output uncertainty estimates, since obtaining meaningful or accurate estimates at the level of individuals is usually computationally and statistically infeasible.} \citep{kim2019multiaccuracy,hebert2018multicalibration}.
We show that when the predictor is multiaccurate (or multicalibrated), then the ranking distribution output by UA ranking satisfies a natural generalization of multiaccuracy (resp. multicalibration) towards the same groups.
This result can be interpreted as an interpolation between individual and group fairness notions for ranking: as the set of subgroups the predictor is multicalibrated against becomes more refined, the UA ranking for predictions more accurately reflects the UA ranking induced by the unknown ground truth classes of individuals.

To investigate the tradeoff between fairness, stability, and utility, in \cref{sec:utility}, we introduce a standard ranking utility model, and show that the utility optimal ranking function cannot hope to achieve stability or fairness guarantees similar to UA.
We also investigate a ranking function which provides a guaranteed tradeoff between stability/fairness and utility. We believe that this will be useful to practitioners interested in employing stable rankings in practice.
Finally, in \cref{sec:experiments}, we corroborate our theoretical results with experimental evidence.

While various notions of stability in rankings have been proposed before (see, e.g., \citet{Asudeh2018StableRankings}), our framework is unique in that it frames the rankings as induced by \emph{predictions} of some machine learning algorithm --- this ties our work more closely to modern applications.
Another benefit of our definition of stability is that it makes progress towards the broader goal of rankings which compose with fair predictors.

\subsection{Related Work}
\label{sec:related-work}
\paragraph{Fairness in Ranking.}
By far the most relevant related work is of \citet{dwork2019learning},
who are also interested in fair rankings induced by predictors, but importantly restrict their focus to only deterministic rankings (where a better prediction means that an individual will always receive a higher rank) induced by probabilistic binary predictors.
Indeed, their motivating example is a setting in which small perturbations to a predictor can massively impact an induced ranking.
By requiring stability of ranking functions, we approach this problem fundamentally differently: we allow (and indeed, require) non-deterministic rankings.
The multiaccuracy and multicalibration guarantees of \citet{dwork2019learning} for induced rankings from predictors are similar in flavor to ours; however, a fundamental difference is that we show this guarantee to be compatible with stability, and, furthermore, that our guarantees hold for each position $k$ in the ranking.

At the intersection of group and individually fair rankings, the work of \citet{Gorantla2023SamplingInd} is most similar to ours.
They show that one can sample from a distribution over rankings which is simultaneously individually and group fair (in a proportional representation sense) for laminar groups.
In contrast, our group fairness hinges on the group-level statistical constraints of multicalibration imposed on the underlying predictor, which instead allow for potentially arbitrarily overlapping groups.
\citet{garcia2021maxmin} also work at the intersection of group and individual fairness in rankings, although their group fairness constraints require that certain groups get representation amongst the top-$k$ positions in the ranking, for all $k \in [n]$.
Both of these works and ours more broadly explore the interplay between group and individual fairness constraints. 

There is far too rich a literature on group and individually fair rankings to cover here, so we restrict attention to only works related to uncertainty and fairness; for a more comprehensive overview, the interested reader is referred to the survey of \citet{zehlike2021fairranking}.

\paragraph{Uncertainty in Rankings.} \citet{rastogi2023fair} investigate fairness in uncertainty aware rankings when the uncertainty estimates themselves may be biased for different subgroups.
We work in the simpler setting in which we assume that uncertainty estimates are themselves unbiased.
\citet{mehrotra2021mitigating} and \citet{mehrotra2022fair} investigate uncertain protected attributes in the settings of subset selection and ranking, respectively.
We do not assume that anything is known about individuals' protected attributes; instead, we only require utilizing the output of a group-fair (multiaccurate) predictor in \cref{sec:multicalibration-guarantees}. Training such a predictor, however, will require certain knowledge of protected attributes (see, e.g., \citet{kim2019multiaccuracy}).

Independently of the line of work on UA rankings \citep{devic23Fairness, singh2021fairness}, \citet{shen2023fairness} propose ranked proportionality, which shares a similar definition.
Their work is in the more general setting of the assignment problem with uncertain priorities, and they focus on algorithmic approaches for achieving a variety of fairness notions simultaneously.
\citet{tang2023learning} also consider the (fair) assignment problem and its connections with calibration.
Our work is instead focused on proving certain \emph{properties} of rankings induced by predictors (predictors which, when stated in the language of \citet{shen2023fairness}, may induce uncertain priorities).
More generally in fairness in uncertain decision making, \citet{tahir2023fairness} consider how different sources of uncertainty can impact fairness.
\citet{guo2023fair} utilize conformal prediction techniques to (feasibly) train fair learn-to-rank models, and are also partially interested in a similar notion of stability as ours.

\citet{guiver2008learning, soliman2009ranking, yang2022can, cohen2021not, penha2021calibration} all also work in the area of ranking with uncertain scores or preferences. In contrast to these works, we simultaneously consider uncertainty, fairness, and stability of rankings.
\citet{heuss2023predictive} also model uncertainty with a Bayesian framework that allows them to apply their method post-hoc to arbitrary retrieval models in hopes of reducing bias.
Perhaps most relevant is the work of \citet{yang2023marginal}, who examine rankings, utility, fairness, and uncertainty simultaneously. They find that modeling uncertainty can actually \emph{improve} utility in some cases, relative to other fair ranking metrics.

\paragraph{Calibration and Ranking.} In \cref{sec:multicalibration-guarantees}, we work with (multi)-calibrated predictors. 
Within the ranking community, there has been some investigation into the impact of calibration of ranking models.
\citet{menon2012predicting} initiated this study, attempting to obtain predicted probabilities based on the score output of a ranking model. \citet{kweon2022obtaining} work in a similar setting, but refine the method of obtaining predicted probabilities.
\citet{yan2022scale} work in the \emph{score-and-sort} model where a scoring function is learned to score each individual, and a ranking function is derived by sorting the individuals according to their scores.
\citet{yan2022scale} aim to ensure that the scoring model is calibrated with respect to some external property.
These works all attempt to infer uncertainty from the scoring function, whereas we assume that uncertainty is given in the form of machine-learned predictions.
\citet{busa2011ranking} show that calibration for ranking functions can help increase diversity of rankings.
More recently, \citet{diciccio2023detection} show that conditional predictive parity (a notion which appears to be related to multicalibration) can help decrease bias in rankings.
These works all highlight the benefits of using calibrated predictive models for ranking, outside of the guarantees that we provide.
\citet{korevaar2023matched} relate calibration and exposure in rankings by comparing the rankings attained by subgroups with similar score distributions.

\paragraph{Stability in Rankings.} In the information retrieval literature, \citet{Asudeh2018StableRankings} also study the notion of stability for rankings.
They work in a setting where a \emph{score} is calculated based on a weighted sum of features of each item, and stability is then with respect to small changes of these weights.
However, their notion of stability is based on geometric intuition for their scoring function and its dual, and only holds for any fixed data set.
They furthermore state that stability is not a property of their ``scoring function'' (particular weighted sum over features).
In contrast, we are explicitly defining stability as a property of our ranking function, which maps from any set of predictions (data set) to a randomized ranking.
\citet{oh2022rank} also study the sensitivity of rankings; however, their context is slightly different: they examine stability with respect to \emph{user interactions} with, e.g., a recommendation system.
In a very recent followup work \citep{oh2024finest}, the same authors also provide an algorithm to empirically achieve stability in that setting.
\citet{bruch2020stochastic} provide experimental evidence showing that randomization can help stability (which they define as \emph{robustness}) during the training of learning-to-rank models. Our theoretical results are complementary and corroborate the empirical evidence of \citet{bruch2020stochastic} that randomized rankings are more robust to noise than deterministic ones.

Finally, in terms of the interplay between prediction systems and rankings, the work of \citet{narasimhan2020pairwise} is perhaps most relevant; they show that the ranking problem can be considered as a pairwise binary classification problem between items to determine which item should be placed at a higher rank.

\section{Notation and Preliminaries}
\label{sec:preliminaries}
We write vectors in boldface.
We use the standard notation $\bm{x}_{-i}$ to denote the vector $\bm{x}$ with the \Kth{i} coordinate removed.
For a random event $\mathcal{E}$, we write $\1_{\mathcal{E}}$ for the indicator function which is 1 if $\mathcal{E}$ happens, and 0 otherwise.
The total variation distance of two measures $\mu, \nu$ is defined as the maximum difference in probability for any event $\mathcal{E}$ under the two measures, i.e., $\dTV{\mu}{\nu} := \max_{\mathcal{E}} |\mu(\mathcal{E}) - \nu(\mathcal{E})| = \half ||\mu - \nu||_1$.
We will use the \emph{entry-wise} matrix norms\footnote{rather than induced norms, which are typically described by the same notation.} $||M||_1 = \sum_{i,j} |M_{i,j}|$ and $||M||_{\infty} = \max_{i,j} | M_{i,j}|$.

$\X$ denotes a set of \emph{individuals}; it contains humans, ads, service requests, or other entities towards whom fairness is desired.
The elements of $\X$ can be labeled with labels from the label set $[L] = \SET{1, 2, \ldots, L}$.
We work in the multiclass ``ordinal'' classification setting where the labels are sorted from most to least preferred as $L \succ L-1 \succ \cdots \succ 2 \succ 1$.
This notation corresponds with the intuition that possessing a higher merit score/class is valued more by a decision maker.
A common special case is $L=2$, i.e., binary labels, where label $1$ might represent irrelevant/unsuitable, while label $2$ represents relevant/suitable.

\subsection{Predictors}

We focus on predictors in the multiclass setting which output distributions over $L$ labels.
Let $\Delta_L$ denote the set of all distributions on $[L]$.
A \emph{(probabilistic) predictor} $f: \X \to \Delta_L$ is a function mapping data points to distributions over labels.
We let $\bm{p} = f(x)$ denote the vector of probabilities that the predictor $f$ assigns to individual $x \in \X$.
For any class $\ell \in [L]$, $p_\ell$ denotes the probability of that class.
As an example, for a probabilistic binary predictor $f$ in the context of determining whether a candidate is qualified for a job, $p_2$ would capture the probability that the individual $x$ is qualified, while $p_1 = 1 - p_2$ is the probability that $x$ is unqualified.

Rankings involve multiple individuals, and hence multiple predictions.
A \emph{prediction for $n$ individuals} $P$ is an $n \times L$ matrix where each row corresponds to the distribution over labels for a particular individual.
We define $\P_{n,L}$ to be the set of all predictions for $n$ individuals, i.e., the set of all $n \times L$ matrices where each row is a distribution.
We will frequently consider the case in which a predictor $f$ for single individuals is applied to each of $n$ individuals separately. 
For a vector $\bm{x} = (x_1, \ldots, x_n) \in \X^n$ of $n$ individuals, we write $f(\bm{x}) = (f(x_1), \ldots, f(x_n))$ for the $n\times L$ matrix of predictions for all of the $n$ individuals.

We use the random variable $\lab{x}$ to denote the (random) label of individual $x$; when we specifically consider an individual $x_i$ in a vector of individuals, we abbreviate $\lab{i} := \lab{x_i}$.
We write $\nlab{\ell} = \sum_i \1_{\lambda_i = \ell}$ for the random variable that is the number of individuals with label $\ell$; when we use this notation, the domain of $i$ will be clear from the context. 
We extend this notation to write $\nlab{\geq \ell} = \sum_{\ell'=\ell}^L \nlab{\ell'}$ for the number of individuals with label $\ell$ or better, and similarly for $\nlab{> \ell}$.
We will sometimes restrict the count to individuals in a particular set $S$, and then write 
$\nlab[S]{\ell} = \sum_{i \in S} \1_{\lambda_i = \ell}$, and similarly for the derived notation.
In particular, we use the notation $\nlab[-i]{\ell} = \nlab[{[n]} \setminus \SET{i}]{\ell}$ for the number of individuals other than $i$ with a particular label $\ell$.

\subsection{Rankings and Ranking Functions}

A principal would like to use predictions provided by a predictor to output a (distribution over) rankings.
As examples, consider a site or service such as LinkedIn providing an employer with a ranked list of applicants to interview \citep{geyik2019fairness}, or an online platform deciding on the order in which to display ads or vendors to a visitor.
In these settings, because \emph{attention} is a limited resource, a common approach would have the principal \emph{rank} the items in question based on some function of the predictions.

A \emph{ranking} is a total order on $n$ individuals.
A \emph{randomized} ranking is a distribution over rankings.
Let $\dsmatrix$ denote the set of all $n \times n$ doubly stochastic matrices. 
Each matrix $M \in \dsmatrix$ represents a \emph{randomized ranking}\footnote{More precisely, it represents the \emph{marginal probabilities} of the distribution, which can typically be implemented by many different distributions over rankings. We assume that individuals care only about the probabilities with which they are ranked in each position, in which case marginal distributions sufficiently capture fairness.} over $n$ individuals, where $M_{i,k}$ is the probability with which individual $i$ is ranked in position $k$.
When reasoning about random rankings, we use $\atrank{i}{k}$ to denote the random event that individual $i$ receives position $k$ in the ranking.

We refer to mappings from predictions to (randomized) rankings as ranking functions:
\begin{definition}    
A \emph{ranking function} $r: \P_{n,L} \to \dsmatrix$ maps predictions over $L$ labels on a data set of $n$ individuals to a randomized ranking of those individuals.
\end{definition}
By focusing on ranking functions, we implicitly state that the principal interacts with the data set \emph{only} through the predictions over labels.
That is, we do not consider \textit{listwise} learning-to-rank schemes --- such as \citet{cao2007learning,xu2007adarank} --- in which the principal directly learns a function mapping data sets of individuals' features to rankings.

\subsection{Desiderata of Ranking Functions}

While ranking functions could be very general, there are natural requirements that make them ``reasonable'' to be used.
In particular, we focus on the following basic properties.

\begin{definition}[Anonymity]
\label{def:anonymity}
    A ranking function $r: \P_{n,L} \to \dsmatrix$ is \emph{anonymous} if every permutation $\sigma: [n] \to [n]$ of the predictions for individuals results in an identical permutation of the individuals' ranks.
\end{definition}

Anonymity states that the outcome for an individual depends only on their (and everyone else's) prediction, but not on the index at which the individual appeared in the data set, i.e., on their identity.
As such, it is an essential fairness requirement in virtually all settings.

A second essential property of ranking functions is \emph{stability}: that small changes in the predictions only lead to small changes in the rankings.
\begin{definition}[Stability]
\label{def:stable_ranking}
  Fix $n$ and $L$.
  A ranking function $r: \P_{n,L} \to \dsmatrix$ is \emph{$\gamma$-stable} if $|| r(P) - r(P') ||_{\infty} \leq \gamma \cdot ||P-P'||_1$ for all predictions $P, P' \in \P_{n,L}$.
\end{definition}
This is particularly important when the predictions are the result of ML-based training methods, which will always contain non-trivial amounts of noise. Indeed, the lack of stability is a well-documented and problematic aspect of many ML-systems, and has been shown not only within the fairness literature \citep{cooper2023my}, but has long been a concern for image classification models \citep{goodfellow2015adversarial} and more recently also LLMs \citep{zou2023universal}.

\section{Predictions and Rankings}
\label{sec:ranking_functions}
We first show useful fairness consequences of stability: combining a stable ranking function with an individually fair predictor results in fair ranking outcomes.
We then show that stability and anonymity are fundamentally at odds with determinism: only constant deterministic ranking functions are stable, and no deterministic ranking function is anonymous. This establishes that randomization is inherently necessary for a ranking function to be meaningful, anonymous, and stable.
We then present our adaptation of the UA ranking function of \citet{singh2021fairness}, and show that it is anonymous and stable.

\subsection{Consequences of Stability}
\label{sec:stability-consequences}

Stability implies that small changes in predictions do not change the distribution over rankings much.
This has two immediate but noteworthy consequences: (1) if the predictions are made by an \emph{individually fair} predictor, then similar individuals will be ranked similarly, and (2) as the predictions approach ground truth, the ranking distribution produced by the ranking function approaches the rankings under the ground truth.

To formalize the first claim, we recall the seminal definition of an \emph{individually fair} predictor \citep{dwork2012fairness}. 
This notion assumes a metric $d$ defined on $\X$ capturing a relevant measure of \emph{similarity} between individuals. 
For $\beta>0$, a probabilistic predictor $f$ is \emph{$(\beta, d)$-individually fair} if $||f(x) - f(x')||_1 \leq \beta \cdot d(x,x')$ for all $x, x' \in \X$.
\begin{proposition}
\label{prop:stability-IF-composition}
  Let $f: \X \to \Delta_L$ be a $(\beta, d)$-individually fair predictor, and $r: \P_{n,L} \to \dsmatrix$ an anonymous and $\gamma$-stable ranking function.
  Given a data set of individuals $(x_i)_{i \in [n]}$ and their associated predictions $P = (f(x_i))_{i \in [n]} \in \P_{n,L}$, let $\bm{q}, \bm{q}'$ be the \Kth{i} and \Kth{j} rows of $r(P)$, respectively.
  Then, $\| \bm{q} - \bm{q}' \|_\infty \leq (2 \beta \gamma) \cdot d(x_i, x_j)$.
\end{proposition}

\begin{proof}
    The proof is a straightforward application of $\gamma$-stability with respect to the given prediction matrix $P$ and a matrix $P'$, where $P'$ is exactly $P$ but with rows $i$ and $j$ swapped (requiring the anonymity condition). This, combined with the definition of $(\beta,d)$-individual fairness for $i, j$, completes the proof.
\end{proof}

The result can be interpreted as a composition guarantee for anonymous and stable rankings with individually fair predictors: if $x, x'$ are simultaneously in a data set, the difference in their distributions over rankings can be at most proportional to their dissimilarity under the metric $d$.
Another interpretation is the following: Stability and individual fairness are both Lipschitz conditions, and composition of Lipschitzness implies that an individually fair predictor combined with a stable ranking will induce an individually fair ranking.

Another very straightforward but desirable consequence of stability is obtained by considering one prediction to be ground truth and the other obtained from a learned classifier.

\begin{corollary}    
\label{cor:nature_close}
    Let $f^*: \X \to \Delta_L$ be the ground truth label distribution for individual $x$, and $f: \X \to \Delta_L$ the learned predictor.
    Assume that $f$ is $\epsilon$-accurate, satisfying that $\| f(x) - f^*(x) \|_1 \leq \epsilon$ for all $x\in \X$.
    Then, any $\gamma$-stable ranking function $r$ guarantees that 
    $\| r(f(\bm{x})) - r(f^*(\bm{x})) \|_\infty \leq \gamma \cdot n\epsilon$ for all $\bm{x} \in \X^n$. 
\end{corollary}
Put differently, for any stable ranking function, accurate individual level uncertainty estimates (relative to a ground truth $f^*$) will induce accurate individual level rankings.
Although somewhat obvious, we highlight this property of stability since the ``ground truth'' approach is often a central assumption in the study of machine learned predictors \citep{shalev2014understanding}.

\subsection{Stability and Determinism are Incompatible}
\label{sec:deterministic}

A third property which most rankings used in practice possess, and which is often considered desirable by practitioners, is determinism: that for given inputs, only one ranking (rather than a distribution over rankings) can result. 

\begin{definition}[Determinism]
\label{def:determinism}
   A ranking function $r: \P_{n,L} \to \dsmatrix$ is \emph{deterministic} iff for all $P \in \P_{n,L}$, the resulting distribution over rankings $r(P)$ has only  entries in $\{ 0, 1 \}$. 
\end{definition}

Perhaps the most well-known deterministic ranking function is given by the Probability Ranking Principle (PRP) of \citet{robertson1977probability}.
In the setting with binary predictions, this ranking function sorts individuals by decreasing probability of belonging to class 2, i.e., being qualified.

Naturally, one may ask whether a deterministic ranking function like the PRP can be stable or anonymous.
Unfortunately, neither is possible, as captured by the following.
\begin{proposition}
\label{prop:det-ranking-not-stable}
  No deterministic ranking function $r: \P_{n,L} \to \dsmatrix$ is anonymous. Furthermore, any deterministic and stable ranking function must be \emph{constant}, in the sense that $|\Ima(r)| = 1$, i.e., the ranking function outputs the same ranking for all input predictions.
\end{proposition}
\begin{proof}
To prove the impossibility of anonymity, consider any prediction matrix $P = (\bm{p})_{i \in [n]}$ with identical predictions $\bm{p}$ for each individual (e.g. $\bm{p} = (1, 0, \dots, 0) \in \Delta_L$).
Then, any deterministic ranking function $r$ must order the individuals based only on their indices in $P$, since they all have identical predictions.
For any permutation $\sigma: [n] \to [n]$, let $P_\sigma$ represent applying permutation $\sigma$ to the rows of $P$. 
Since the ranking function can only depend on the input matrix, and $r$ is deterministic, we have that $r(P) = r(P_\sigma)$.
However, by the definition of anonymity (\cref{def:anonymity}), the permutation $\sigma$ on the rows of $P$ should produce the permutation $r(P)_\sigma$ on the individuals in the resulting ranking, which is a contradiction.
Therefore, $r$ is not anonymous.

To prove the instability result, we prove the contrapositive. 
Let $r$ be deterministic and non-constant. We will show that $r$ is not stable.
Because $r$ is non-constant, there exist $P, P' \in \P_{n,L}$ with $r(P) \neq r(P')$. Consider the straight line $Q(\beta) = \beta P + (1-\beta) P'$, for $\beta \in [0,1]$. Because $\P_{n,L}$ is convex, $Q(\beta) \in \P_{n,L}$ for all $\beta \in [0,1]$.
Let $\beta^* = \inf \Set{\beta}{Q(\beta) = P'}$; $\beta^*$ is well-defined because $Q(1) = P'$.
By definition, $Q(\beta) \neq P'$ for all $\beta < \beta^*$, and if $\beta^*=0$, then $Q(\beta^*) = P \neq P'$. 
On the other hand, by the definition of the infimum, for every $\delta > 0$, there is a $\delta' < \delta$ with $Q(\beta^* + \delta') = P'$.
Thus, we obtain arbitrarily close pairs $\beta' \leq \beta^* < \beta''$ with $Q(\beta') \neq Q(\beta'')$.
Because $r$ is deterministic, all entries of $r(Q(\beta'))$ and $r(Q(\beta''))$ are in $\SET{0,1}$, implying that $||r(Q(\beta')) - r(Q(\beta''))||_\infty \geq 1$. 
On the other hand, $||Q(\beta') - Q(\beta'')||_1 \leq ||2\delta (P'-P) ||_1 \to 0$ as $\delta \to 0$.
This implies that $r$ is not stable, completing the proof.
\end{proof}

We remark that the instability portion of the proof did not rely on considering a straight line.
By considering any path (curve in $\mathbb{R}^{n \times L}$) connecting $P, P'$ and its parametrization by $\beta \in [0,1]$, the exact same proof still works. This shows that even if we consider only a subset of possible predictions, so long as the subset is path-connected\footnote{Recall that a set $A$ is path-connected if for every pair of elements $x,y\in A$ there exists a continuous path between $x$ and $y$ which is entirely contained within $A$.}, a deterministic stable ranking function must be constant.
This extends the proposition to settings where prediction strategies may output only certain (path-connected) subsets of predictions, due to, for example, intrinsic preferences or implicit bias of a particular learning algorithm. 

\Cref{prop:det-ranking-not-stable} formalizes the intuition that randomness is required to achieve stability.
Indeed, the main results of our work also show that randomization and the resulting stability are crucial for achieving desirable \emph{fairness} guarantees.

\subsection{Uncertainty Aware Rankings}
\label{sec:up-rankings}
Meaningful deterministic ranking functions cannot be stable; in fact, it is not immediate that there exist (non-constant) stable ranking functions.
We now show that \emph{Uncertainty Aware (UA) Rankings},
introduced by \citet{singh2021fairness}, are anonymous and stable.

UA rankings were originally introduced via an axiomatization of when a probabilistic ranking should be considered ``fair'' for given merit distributions.
\citet{devic23Fairness} further refined this axiomatization by combining notions of meritocracy and \emph{lifting} deterministic decision making to decision making under uncertainty.

The definition of \citet{singh2021fairness} assumed that merit distributions were continuous and ties occurred with probability 0.
Motivated by predictors which output distributions over discrete label sets such as 1--5 or \{irrelevant, suitable, extremely relevant\} with a corresponding total order, we adapt the definition of UA rankings:
\begin{definition}[Uncertainty Awareness \citep{singh2021fairness}]
\label{def:UA}
  A randomized ranking $M \in \dsmatrix$ is \emph{uncertainty aware} for a prediction $P \in \P_{n, L}$ if for each individual $i$ and position $k$, the entry $M_{i,k}$ is the probability that $i$ has the \Kth{k} highest label if all labels $\lambda_i \sim \bm{p}_i$ are sampled independently from the respective distributions $\bm{p}_i \in \Delta_L$, and ties are broken uniformly.
  Formally, conditioned on the drawn labels $\lab{i}$ of all individuals $i$, which entail the counts $\nlab{\ell}$ for all labels, the probability for individual $i$ to obtain rank $k$ is
  \begin{align}
    \ProbC{\atrank{i}{k}}{\lab{i} = \ell, \nlab{1}, \ldots, \nlab{L}}
    = \begin{cases} 
    \frac{1}{\nlab{\ell}} & \text{if } \nlab{> \ell} < k \leq \nlab{\geq \ell} \\
    0 & \text{otherwise}.
    \end{cases}
    \label{eqn:UA-tiebreaking-RV}
  \end{align}
  A ranking function $r: \P_{n, L} \to \dsmatrix$ is uncertainty aware if $r(P)$ is uncertainty aware for all $P \in \P_{n, L}$.
\end{definition}
Because the definition of uncertainty awareness fully prescribes the ranking distribution $M$ for a given prediction $P$ (as shown in Lemma~4.2 of \citet{singh2021fairness}), there is a unique uncertainty aware ranking function $r$ for any given $n, L$; we henceforth denote it by $\RUP$.

Intuitively, the fairness of UA can be interpreted through a \emph{possible futures} viewpoint.
Given two individuals $A$, $B$, if $A$ has more merit than $B$ in 60\% of futures (when the me\-rits/labels of both $A$ and $B$ are sampled from their respective distributions), then UA implements the requirement that the allocation in the present should respect this uncertainty and give $A$ the better rank at least 60\% of the time (and $B$ at least 40\% of the time); this entails the need for randomization.
We refer the reader to \citet{devic23Fairness, singh2021fairness} for a formal argument on the fairness of UA ranking.

Our first key insight for proving properties of UA rankings is that by taking into account the randomness of the draws of labels and the tie breaking, the rank distribution produced by UA ranking can be summarized as follows:
\begin{proposition} \label{prop:UA-summary}
   Let $P \in \P_{n, L}$ be a prediction, and $\rup{P}$ the ranking distribution produced by $\RUP$ for $P$.
   Then, the probability of individual $i \in [n]$ being ranked in position $k \in [n]$ is:
   \begin{align}
       \ProbC[\rup{P}]{\atrank{i}{k}}{\lab{i} = \ell}
       &= \sum_{j=0}^{n-1} \frac{1}{j+1} \cdot \Prob[P]{\nlab[-i]{\ell} = j \text{ and } k-(j+1) \leq \nlab[-i]{> \ell} < k},
       \label{eqn:rank-prob-only-label} \\
       \Prob[\rup{P}]{\atrank{i}{k}} 
        &= \sum_{\ell} p_{i,\ell} \cdot \ProbC[\rup{P}]{\atrank{i}{k}}{\lab{i} = \ell}.
       \label{eqn:rank-total-probability}
    \end{align}
\end{proposition}    

\begin{proof}
    For the first part, we observe that the probability of $\atrank{i}{k}$ in \eqref{eqn:UA-tiebreaking-RV} depends only on $\nlab{\ell}$ and $\nlab{> \ell}$.
    By considering all the possible values of $N^{\geq \ell}$ for which \eqref{eqn:UA-tiebreaking-RV} gives a non-zero probability, we obtain that 
    \begin{align*}
       \ProbC[\rup{P}]{\atrank{i}{k}}{\lab{i} = \ell}
        & = \sum_{j=1}^n \frac{1}{j} \cdot \ProbC[P]{\nlab{\ell} = j \text{ and } \nlab{> \ell} < k \leq \nlab{\geq \ell}}{\lab{i} = \ell}.
    \end{align*}
    The result is then obtained by noticing that conditioned on $\lab{i} = \ell$, we have $\nlab[-i]{\ell} = \nlab{\ell} - 1$ and $\nlab[-i]{\ell'} = \nlab{\ell'}$ for all $\ell' \neq \ell$.
    The second part of the proposition simply states the law of total probability.
\end{proof}

We also remark that the definition of \citet{singh2021fairness}, in contrast to \cref{def:UA}, did not require the labels/merits of different individuals to be sampled \emph{independently} from their respective distributions $P_i$. We add this independence requirement to facilitate connections with learning algorithms for predictors.
An added benefit of the independence assumption is that it makes it possible to explicitly compute $\rup{P}$ in polynomial time, as captured by the following proposition. 
Note that this is in contrast to the case of possibly correlated labels/merits from previous work \citep{singh2021fairness, devic23Fairness};
indeed, a main technical contribution of these works was analyzing the loss in fairness/utility incurred due to imperfectly approximating $\rup{P}$ via sampling.

\begin{proposition}
\label{prop:exact-ranking}
   There exists an algorithm which, given $P \in \P_{n, L}$, exactly computes $\rup{P}$ in time $O(n^4 + n^3 L)$.
\end{proposition}

\begin{proof}
   The two parts of \cref{prop:UA-summary} combined imply that in order to compute row $i$ of $\rup{P}$, it is sufficient to compute $\Prob[P]{\nlab[-i]{\ell} = j \text{ and } k-(j+1) \leq \nlab[-i]{> \ell} < k}$ for all pairs $(j, k) \in [m]$.
   This is accomplished by a dynamic program similar to a standard undergraduate exercise, which is to compute a Poisson Binomial distribution explicitly.

   For notational convenience, assume that $i=n$; this is solely to avoid a special case in the recurrence, and also without loss of generality by anonymity of the UA rule.
   For any $t, j, j' \in \SET{0, 1, \ldots, m-1}$, let 
   \begin{align*}
       A(t, j, j') & = \Prob[P]{\nlab[\SET{1, \ldots, t}]{\ell} = j \text{ and } \nlab[\SET{1, \ldots, t}]{> \ell} = j'}
   \end{align*} be the probability that among the first $t$ individuals, exactly $j$ have label $\ell$, and exactly $j'$ have a label strictly better than $\ell$.
   From these values, we can then construct the necessary quantities as 
   \begin{align*}
   \Prob[P]{\nlab[-n]{\ell} = j \text{ and } k-(j+1) \leq \nlab[-n]{> \ell} < k}
   & = \sum_{j'=k-(j+1)}^{k-1} A(n-1, j, j').
   \end{align*}

   We give the recurrence relationship for the $A(t,j,j')$. 
   The base cases are that
   \begin{align*}
   A(0,0,0) & = 1
   \\ A(0,j,j') & = 0 \text{ if } j + j' \neq 0,
   \end{align*}
   because with no individuals, the only possible numbers of individuals with given labels is 0.

   Now consider $A(t,j,j')$ for $t \geq 1$.
   With probability $p_{t,\ell}$, individual $t$ has label $\ell$, in which case the desired event happens when $j-1$ individuals among the first $t-1$ have label $\ell$, and $j'$ have labels strictly better than $\ell$.
   With probability $\sum_{\ell' > \ell} p_{t,\ell'}$, individual $t$ has a label strictly better than $\ell$, in which case the desired event happens when $j$ individuals among the first $t-1$ have label $\ell$, and $j'-1$ have labels strictly better than $\ell$.
   Finally, with probability $\sum_{\ell' < \ell} p_{t,\ell'}$, individual $t$ has a label strictly worse than $\ell$, in which case the desired event happens when $j$ individuals among the first $t-1$ have label $\ell$, and $j'$ have labels strictly better than $\ell$.
   These three cases disjointly cover all possibilities for the label of $t$, so we have derived the following recurrence:

   \begin{align*}
       A(t,j,j') 
       & = p_{t,\ell} \cdot A(t-1,j-1,j')
       + \sum_{\ell' > \ell} p_{t,\ell'} \cdot A(t-1,j,j'-1)
       + \sum_{\ell' < \ell} p_{t,\ell'} \cdot A(t-1,j,j').
   \end{align*}

   Here, to avoid case distinctions for whether $j$ and/or $j'$ are 0, we treat $A(t,j,j') = 0$ whenever $j$ or $j'$ are negative.

   Notice that for any fixed $t$, all values $\sum_{\ell' > \ell} p_{t,\ell'}$, being prefix sums, can be pre-computed in time $O(L)$.
   Thus, for all $\ell, t$, the precomputation can be performed in time $O(nL)$.

   Then, any one entry $A(t,j,j')$ can be computed in constant time from previously computed values. Because the table has size $O(n^3)$, the total computation takes time $O(n^3)$.
   Summing over all possible values of $i$, the total time to compute the entire ranking distribution $\rup{P}$ is $O(n^4 + n^2 L)$.
   Finally, the post-processing of computing the $\Prob[P]{\nlab[-i]{\ell} = j \text{ and } k-(j+1) \leq \nlab[-i]{> \ell} < k}$ for all $k$, for fixed $i, \ell, j$, can be implemented in time $O(n)$ by using differences of prefix sums; thus, all values can be computed in time $O(n^3 L)$.
   This gives a total time of $O(n^4 + n^3 L)$.
\end{proof}

While the notion and use of uncertainty aware rankings may appear to be of primarily theoretical interest, it is in fact used in practice. For example, the NBA draft lottery can be understood through the lens of uncertainty aware rankings. The \emph{merit} of a team is its need for better choice picks, which can be (imperfectly) inferred from the team's performance in the previous season. The draft order is then obtained by a weighted lottery based on these uncertain merits.

We now present the central result of this section: that the uncertainty aware ranking function is anonymous and stable.
\begin{theorem}
\label{prop:stability_of_upr}
Let $\RUP: \P_{n, L} \to \dsmatrix$ be the UA ranking function for $n$ individuals and $L$ labels. 
$\RUP$ is anonymous and $1$-stable.
\end{theorem}

The following lemma is a key part of the proof of stability; it bounds how different the probabilities for individual $i$ obtaining rank $k$ can be under two different prediction matrices, as a function of how similar these matrices are:
\begin{lemma} \label{lem:rank-bound}
   Let $P, Q \in \P_{n, L}$ be two different prediction matrices.
   For any individual $i$, let $\bm{p}_{i}, \bm{q}_{i}$ be the \Kth{i} rows of $P, Q$, respectively, i.e., the label distributions of individual $i$ under the two predictions.
   Let $i$ be an individual, $k \in [n]$ a position, and $\ell \in [L]$ a label.
   Then,
   \begin{align*}
       \left| \ProbC[\rup{P}]{\atrank{i}{k}}{\lab{i} = \ell} - \ProbC[\rup{Q}]{\atrank{i}{k}}{\lab{i} = \ell}\right|
      & \leq 
      2 \cdot \sum_{i' \neq i} \dTV{\bm{p}_{i'}}{\bm{q}_{i'}}.
   \end{align*}
\end{lemma}
We defer the proof of \cref{lem:rank-bound} to \cref{appx:ommited-proofs}.

\begin{proof}[Proof of \cref{prop:stability_of_upr}]
    First, the UA ranking rule is obviously anonymous, simply by its (symmetric) definition which treats all indices identically. Thus, we focus on proving stability.

    Now, let any individual $i$ and rank $k$ be given.
    By \cref{eqn:rank-total-probability} in the second part of \cref{prop:UA-summary}, 
    $\Prob{\atrank{i}{k}} = \sum_\ell p_{i,\ell} \cdot \ProbC{\atrank{i}{k}}{\lab{i} = \ell}$.
    Now consider two different predictors $P, Q$. We bound the difference in probabilities for $i$ to be ranked in position $k$ as follows:

    \begin{align}
    &\left| \Prob[\rup{P}]{\atrank{i}{k}} - \Prob[\rup{Q}]{\atrank{i}{k}} \right| \nonumber \\
     &\qquad \leq 
       \sum_\ell \left| p_{i,\ell} \cdot \ProbC[\rup{P}]{\atrank{i}{k}}{\lab{i} = \ell} - q_{i,\ell} \cdot \ProbC[\rup{Q}]{\atrank{i}{k}}{\lab{i} = \ell} \right| \nonumber
    \\ &\qquad \leq 
       \sum_\ell | p_{i,\ell} - q_{i, \ell} | \cdot \ProbC[\rup{P}]{\atrank{i}{k}}{\lab{i} = \ell} 
        + q_{i,\ell} \cdot \left| \ProbC[\rup{P}]{\atrank{i}{k}}{\lab{i} = \ell} - \ProbC[\rup{Q}]{\atrank{i}{k}}{\lab{i} = \ell} \right| \nonumber
     \\ &\qquad \leq 
       \sum_\ell | p_{i,\ell} - q_{i,\ell} | + q_{i,\ell} \cdot 
       \left| \ProbC[\rup{P}]{\atrank{i}{k}}{\lab{i} = \ell} - \ProbC[\rup{Q}]{\atrank{i}{k}}{\lab{i} = \ell}\right|. 
       \label{eqn:difference-with-p-q}
    \end{align}
    
   By \cref{lem:rank-bound}, we can bound
   $
       \left| \ProbC[\rup{P}]{\atrank{i}{k}}{\lab{i} = \ell} - \ProbC[\rup{Q}]{\atrank{i}{k}}{\lab{i} = \ell}\right|
      \leq 
      2 \cdot \sum_{i' \neq i} \dTV{\bm{p}_{i'}}{\bm{q}_{i'}}
      $.

  Substituting this bound back into \eqref{eqn:difference-with-p-q}, we now obtain that
  \begin{align*}
    | \Prob[\rup{P}]{\atrank{i}{k}} - \Prob[\rup{Q}]{\atrank{i}{k}} |
    & \leq \sum_\ell | p_{i,\ell} - q_{i,\ell} | + 
    \sum_\ell q_{i,\ell} \cdot 
    2 \cdot \sum_{i' \neq i} \dTV{\bm{p}_{i'}}{\bm{q}_{i'}}
    \\ & = \half \dTV{\bm{p}_i}{\bm{q}_i} + 
    2 \cdot \sum_{i' \neq i} \dTV{\bm{p}_{i'}}{\bm{q}_{i'}}
    \\ & \leq ||P-Q||_1.
  \end{align*}

   In the final step, we absorbed the total variation distance for $i$ into the sum for $i'\neq i$ (which has a larger coefficient), and used that the 1-norm is exactly twice the total variation distance.
\end{proof}

As the number of labels increases, the predictor can provide the ranking function with more fine-grained information, which should allow the UA ranking function to produce a wider class of distributions over rankings. 
The next proposition shows that this is indeed the case:
\begin{proposition}
\label{prop:multiclass_more_expressive}
  The expressivity of uncertainty aware ranking functions is strictly increasing in $L$.
  More formally, let $n \geq L$, and 
  $\RUP[L]: \P_{n, L} \to \dsmatrix,      \RUP[L-1]: \P_{n, (L-1)} \to \dsmatrix$ be the corresponding UA ranking functions.
  Then, $\rup[L]{\P_{n, L}} \supsetneq \rup[L-1]{\P_{n, (L-1)}}$.
\end{proposition}
\begin{proof}
    First, to see monotonicity, notice that adding a column of all 0 entries, i.e., an unused label $L$, does not change the behavior of $\RUP$. For any $P \in \P_{n, (L-1)}$, writing $[P, \bm{0}] \in \P_{n, L}$ for this matrix, we have $\rup[L]{[P, \bm{0}]} = \rup[L-1]{P}$, implying that $\rup[L]{\P_{n, L}} \supseteq \rup[L-1]{\P_{n, L-1}}$. 
  
    To prove strictness of inclusion, consider a prediction $P=J_n$ over $n=L$ individuals. Here, $J_n$ is the $n \times n$ row-reversed identity matrix with ones along the anti-diagonal, so individual $i$ is deterministically known to have label $L-i+1$. Then $\rup{P} = I_n$ for the $n\times n$ identity matrix $I_n$, i.e., individual $i$ is ranked deterministically in position $i$, and we have proved that $I_n \in \rup[L]{\P_{n, L}}$.
    Note that due to the tie breaking of $\RUP$, to achieve a deterministic ranking, no two individuals must ever have the same label, i.e., the supports of the $n=L$ rows of any prediction matrix $Q$ yielding $\rup{Q} = I_n$ must be disjoint. This implies that $Q$ must have at least $L$ columns, i.e., $I_n \notin \rup[L-1]{\P_{n, L-1}}$, completing the proof of strictness of inclusion.    
\end{proof}

Using \cref{prop:multiclass_more_expressive}, we can show that our stability analysis is essentially tight up to a factor of 2, ruling out the possibility of, for example, $\tfrac{1}{n}$-stability for UA rankings:
\begin{proposition}
\label{prop:stab-lb}
   For any $L \geq 3$ and $n \geq 2$, $\rua$ is not $\gamma$-stable for any $\gamma < \tfrac{1}{2}$.
\end{proposition}
\begin{proof}
    Let $n\geq 2$ be given.
    We only consider $L = 3$; for any $L > 3$, it suffices by \cref{prop:multiclass_more_expressive} to embed the following instance and ignore the extra labels.
    Consider the prediction matrix $P$ with individual 1 having prediction $\bm{p}_1 = (1/2, 0, 1/2)$, and individuals 2 through $n$ having prediction $\bm{p}_2 = (0, 1, 0)$. 
    Similarly, the prediction matrix $P'$ will have individual 1 with prediction $\bm{p}'_1 = (1,0,0)$ and individuals 2 through $n$ with prediction $\bm{p}_2$.
    That is, $P$ and $P'$ are identical except for individual 1.

    Let $M = \rua(P), M' = \rua(P')$ be the resulting probabilities for placing individuals in specific positions.
    Since all individuals except individual 1 have deterministic qualifications, under $P$, there is a 50\% probability that individual 1 is ranked last, so $M_{1,n} = 1/2$, whereas $M'_{1,n}=0$.
    Therefore, we have the following.
    \begin{align*}
        \| \rua(P) - \rua(P') \|_\infty & \geq |M_{1,n} - M'_{1,n} | = 1/2,
        & 
        \| P - P' \|_1 & = 1.
    \end{align*}
    Thus, $\rua$ cannot be $\gamma$-stable for any $\gamma < 1/2$.
\end{proof}

\section{Multigroup Fairness Guarantees}
\label{sec:multicalibration-guarantees}
We now present our main result: UA rankings naturally compose with multiaccurate and multicalibrated predictors.

We have shown that UA rankings are stable, and, furthermore, that stable rankings compose harmoniously with individually fair predictors.
\cref{cor:nature_close} demonstrates that an accurate predictor, at the individual level, can combine with a stable ranking function (such as UA) to induce a ranking which is close to that of the underlying ground truth.
In practice, however, obtaining accurate uncertainty estimates at the individual level is too strong of an assumption for arbitrary data sets of individuals. This is because such a requirement is equivalent to learning the \emph{Bayes optimal predictor} (the true distribution over the labels conditioned on the features of an individual) \citep[Section~3.2.1]{shalev2014understanding}, which generally requires the number of samples or running time to be exponential in the dimensionality of the features used for prediction, which can be statistically or computationally infeasible.

Instead, we focus on obtaining a coarser guarantee for UA rankings at the level of \emph{subgroups} of the domain for data sets \emph{sampled} i.i.d.~from a distribution $\D$ over individuals (instead of arbitrary data sets).
The i.i.d.~assumption is a standard setting for machine learning, and has proven useful in many practical settings.
Our contributions are tightly connected to the statistical group-fairness conditions of \emph{multiaccuracy} and \emph{multicalibration} \citep{kim2019multiaccuracy, hebert2018multicalibration}.
Our guarantees will be meaningful since they directly imply that, relative to an underlying ground truth, unbiased predictors will induce unbiased rankings.

\subsection{Group-wise Accuracy Guarantees}
We first recall the definitions of multiaccuracy and multicalibration from the fair machine learning literature.
Then, we state our result on the average accuracy of rankings induced by multiaccurate and multicalibrated predictors when compared to rankings induced from nature.

\begin{definition}[Multiaccuracy/Multicalibration \citep{kim2019multiaccuracy, hebert2018multicalibration}]
  Let $\mathcal{D}$ be a distribution over individuals $\mathcal{X}$.
  Let $f^*$ be the ground truth distribution of labels, i.e., $f^*(x)$ is the true distribution of labels for individual $x$, while $f$ is the predictor, so $f(x)$ is the predicted label distribution.
  
  Let $\mathcal{C}$ be a collection of sets for which the predictor is to be multiaccurate/multcalibrated.
  Let $\alpha \geq 0$ be a parameter for how far from fully accurate/calibrated the predictor is allowed to be.
  When writing $\Expect{\bm{v}}$ for a vector-valued quantity $\bm{v}$, we mean the coordinate-wise expectations.
  Then we have that:
  \begin{enumerate}
    \item $f$ is \emph{$(\mathcal{C}, \alpha)$-multiaccurate} if for every set $S \in \mathcal{C}$,
      \[
      \| \Expect[x \sim \mathcal{D}]{\1_{x \in S} \cdot (f^*(x) - f(x))} \|_{\infty}
      \leq \alpha.
      \]
      That is, for each of the given groups $S \in \mathcal{C}$ and each label, the expected probability mass on that label is approximately the same for the predictor as for the ground truth.

    \item Let some interval width $\delta\in(0,1]$ be given, such that $1/\delta$ is an integer. 
      $f$ is \emph{$(\mathcal{C}, \alpha, \delta)$-multicalibrated} if for every set $S \in \mathcal{C}$ and vector $(j_1, j_2, \ldots, j_{L}) \in \SET{0, 1, \ldots, 1/\delta-1}^L$,
      \[
      \| \Expect[x \sim \mathcal{D}]{\1_{x \in S} \cdot \1_{\forall \ell,\ f(x)_{\ell} \in [j_{\ell} \cdot \delta, (j_{\ell}+1) \cdot \delta)} \cdot (f^*(x) - f(x))} \|_{\infty} \leq \alpha.
      \]
      That is, in addition to fixing a group, even if we also fix a (rough) interval within which the predicted probability mass must lie, the predictor still has to be close to the ground truth, for each possible label.
  \end{enumerate}
\end{definition}
Each set $S \in \mathcal{C}$ represents a protected group of import in the underlying population $\X$.
The sets in $\mathcal{C}$ can be complex, overlapping, nested, laminar, etc., since both multiaccuracy and multicalibration with respect to $\mathcal{C}$ are --- in contrast to other notions of group fairness in supervised learning such as equalized odds \citep{awasthi2020equalized} or equality of opportunity \citep{hardt2016equality} --- statistically sound in that they are \emph{consistent} with the underlying ground truth $f^*$.
There are a variety of learning and post-processing algorithms which, in terms of sample and time complexity, efficiently achieve multiaccuracy/multicalibration \citep{hebert2018multicalibration, kim2019multiaccuracy, gopalan2022low, haghtalab2023unifying}.

We now present the central contribution of this section: multicalibration and multiaccuracy for a predictor $f$ guarantee that on a per group basis, UA rankings derived from $f$ will be close to UA rankings derived from the ground truth $f^*$.
\begin{theorem} \label{prop:multiX-bound}
  Let $\mathcal{D}$ be a distribution over individuals $\mathcal{X}$.
  Let $f^*$ be the ground truth distribution of labels, and $f$ a predictor.
  For any $n$, let $\mathcal{D}^n$ be the distribution obtained from drawing a vector of $n$ i.i.d.~samples from $\mathcal{D}$.
  
  Let $\mathcal{C}$ be a collection of sets with $\mathcal{X} \in \mathcal{C}$, the sets of individuals for which the predictor $f$ will be assumed to be multiaccurate/multcalibrated.
  Let $\alpha \geq 0$ be a parameter for how far from fully accurate/calibrated the predictor is allowed to be.
  Then, we have that:
  \begin{enumerate}
     \item If $f$ is $(\mathcal{C}, \alpha)$-multiaccurate, then the following holds for all sets $S \in \mathcal{C}$ and $k \in [n]$: 
        \begin{align*}
            \left| 
                \Expect[\bm{x} \sim \mathcal{D}^n, i \sim \text{Unif}({[n]})]{\1_{x_i \in S} \cdot \left( 
                \Prob[\rup{f^*(\bm{x})}]{\atrank{i}{k}} - \Prob[\rup{f(\bm{x})}]{\atrank{i}{k}}
                \right) }
            \right|
            \leq L n \alpha.
        \end{align*}
    \item Let some interval width $\delta \in (0,1]$ be given, such that $1/\delta$ is an integer. 
    If $f$ is $(\mathcal{C}, \alpha, \delta)$-multicalibrated and $(\SET{\X}, \alpha)$-multiaccurate, then for every set $S \in \mathcal{C}$, vector $(j_1, j_2, \ldots, j_{L}) \in \SET{0, 1, \ldots, 1/\delta-1}^L$, and $k \in [n]$:
    \begin{align*}
       \Big| 
            \mathbb{E}_{\bm{x} \sim \mathcal{D}^n, i \sim \text{Unif}({[n]})} \Big[
                \1_{x_i \in S} \cdot \1_{f(x_i)_{\ell} \in [j_{\ell} \cdot \delta, (j_{\ell}+1) \cdot \delta) \text{ for all } \ell}  \cdot 
                \left( 
                \Prob[\rup{f^*(\bm{x})}]{\atrank{i}{k}} - \Prob[\rup{f(\bm{x})}]{\atrank{i}{k}}
                \right) \Big]
        \Big|
        \leq L n \alpha.
    \end{align*}
  \end{enumerate}
\end{theorem}

\begin{proof}
   Fix a set $S \in \mathcal{C}$ and position $k \in [n]$. 
   The proofs of both parts of the theorem are essentially identical, with only minor (practically syntactic) changes, and we will give both proofs simultaneously, pointing out the few places where differences occur.
   For the proof of the second part of the theorem (for the case of a multi-calibrated predictor), let $\delta \in (0,1]$ be given, such that $1/\delta$ is an integer, and fix a vector $(j_1, j_2, \ldots, j_{L}) \in \SET{0, 1, \ldots, 1/\delta-1}^L$.
   We use the notation $\one{x}$ to denote $\one{x} := \1_{x \in S}$ for the proof under a multiaccurate predictor $f$, and to denote $\one{x} := \1_{x \in S} \cdot \1_{f(x)_{\ell} \in [j_{\ell} \cdot \delta, (j_{\ell}+1) \cdot \delta) \text{ for all } \ell}$ for the proof under a multicalibrated predictor $f$.

   First, we observe that we can simplify notation by using that all draws of the $x_j$ are i.i.d., and we can apply the law of total probability:
   \begin{align*}
        \\ & \left| \Expect[\bm{x} \sim \D^n, i \sim \text{Unif}({[n]})]{\one{x_i} \cdot 
        \left( \Prob[\rup{f^*(\bm{x})}]{\atrank{i}{k}} - \Prob[\rup{f(\bm{x})}]{\atrank{i}{k}} \right) } \right|
        \\ & \qquad = \left| \Expect[\bm{x} \sim \D^n]{\one{x_n} \cdot
        \left( \Prob[\rup{f^*(\bm{x})}]{\atrank{n}{k}} - \Prob[\rup{f(\bm{x})}]{\atrank{n}{k}} \right) } \right|
        \\ & \qquad = \left| \Expect[x_n \sim \D]{\one{x_n} \cdot \Expect[\bm{x}_{-n} \sim \D^{n-1}]{
        \Prob[\rup{f^*(\bm{x}_{-n},x_n)}]{\atrank{n}{k}} - \Prob[\rup{f(\bm{x}_{-n},x_n)}]{\atrank{n}{k}}
        }} \right|.
   \end{align*}
   
   Next, we sum over all possible labels $\lab{n} = \ell$, condition on those labels for individual $n$, and then omit from the probabilities those random variables that do not affect the probability, to rewrite the preceding expression as
   
   \begin{align}
        & \Bigg| \mathbb{E}_{x_n \sim \D} \Bigg[ \one{x_n} \cdot \mathbb{E}_{\bm{x}_{-n} \sim \D^{n-1}} \Bigg[
        \sum_{\ell}
        \Prob[f^*(\bm{x}_{-n},x_n)]{\lab{n} = \ell} \cdot \ProbC[\rup{f^*(\bm{x}_{-n},x_n)}]{\atrank{n}{k}}{\lab{n} = \ell} \nonumber
        \\& \qquad - \Prob[f(\bm{x}_{-n},x_n)]{\lab{n} = \ell} \cdot \ProbC[\rup{f(\bm{x}_{-n},x_n)}]{\atrank{n}{k}}{\lab{n} = \ell} 
        \Bigg] \Bigg] \Bigg| \nonumber
        \\ & = \Bigg| \mathbb{E}_{x_n \sim \D} \Bigg[ \one{x_n} \cdot  \mathbb{E}_{\bm{x}_{-n} \sim \D^{n-1}} \Bigg[
        \sum_{\ell}
        \Prob[f^*(x_n)]{\lab{n} = \ell} \cdot \ProbC[\rup{f^*(\bm{x}_{-n})}]{\atrank{n}{k}}{\lab{n} = \ell} \nonumber
        \\& \qquad - \Prob[f(x_n)]{\lab{n} = \ell} \cdot \ProbC[\rup{f(\bm{x}_{-n})}]{\atrank{n}{k}}{\lab{n} = \ell} 
        \Bigg] \Bigg] \Bigg| \nonumber
        \\ & \leq \left| \Expect[x_n \sim \D]{\one{x_n} \cdot  \Expect[\bm{x}_{-n} \sim \D^{n-1}]{ 
        \sum_{\ell}
        \left( \Prob[f^*(x_n)]{\lab{n} = \ell} - \Prob[f(x_n)]{\lab{n} = \ell} \right)
        \cdot \ProbC[\rup{f^*(\bm{x}_{-n})}]{\atrank{n}{k}}{\lab{n} = \ell}}} \right| \nonumber
        \\ & \qquad + \left| \Expect[x_n \sim \D]{\one{x_n} \cdot  \Expect[\bm{x}_{-n} \sim \D^{n-1}]{ 
        \sum_{\ell}
        \Prob[f(x_n)]{\lab{n} = \ell} \cdot 
        \left( \ProbC[\rup{f^*(\bm{x}_{-n})}]{\atrank{n}{k}}{\lab{n} = \ell} 
        - \ProbC[\rup{f(\bm{x}_{-n})}]{\atrank{n}{k}}{\lab{n} = \ell} \right)
        }} \right| \nonumber
      \\ & = \left| \sum_{\ell}
        \left( \Expect[x_n \sim \D]{\one{x_n}
        \cdot \left( \Prob[f^*(x_n)]{\lab{n} = \ell} - \Prob[f(x_n)]{\lab{n} = \ell} \right)} \right)
        \cdot \Expect[\bm{x}_{-n} \sim \D^{n-1}]{ 
        \ProbC[\rup{f^*(\bm{x}_{-n})}]{\atrank{n}{k}}{\lab{n} = \ell}
        } \right| \label{eq:term_1}
        \\ & \qquad + \left| \sum_{\ell} 
        \Expect[x_n \sim \D]{\one{x_n} \cdot  
        \Prob[f(x_n)]{\lab{n} = \ell}} \cdot
        \Expect[\bm{x}_{-n} \sim \D^{n-1}]{ 
        \ProbC[\rup{f^*(\bm{x}_{-n})}]{\atrank{n}{k}}{\lab{n} = \ell} 
        - \ProbC[\rup{f(\bm{x}_{-n})}]{\atrank{n}{k}}{\lab{n} = \ell}
        } \right| \label{eq:term_2}.
   \end{align}

   We bound the two sums separately.
   For the first sum, we simply bound   
   \begin{align*}
      \eqref{eq:term_1} 
      \leq \sum_{\ell}
        \left| \Expect[x_n \sim \D]{\one{x_n} 
        \cdot \left( \Prob[f^*(x_n)]{\lab{n} = \ell} - \Prob[f(x_n)]{\lab{n} = \ell} \right) } \right|
      \leq L \cdot \alpha.
   \end{align*}
   In the final step, we applied the multi-accuracy guarantee for $S$ for every label $\ell$ under the sum for the first part of the theorem, and the multi-calibration guarantee for $S$ and $(j_1, \ldots, j_L)$ for every label $\ell$ for the second part of the theorem.

   For the second term, we apply the triangle inequality, and bound
   \begin{align*}
        &\eqref{eq:term_2} 
        \leq \sum_{\ell} 
        \Expect[x_n \sim \D]{\Prob[f(x_n)]{\lab{n} = \ell}} \cdot
        \left|
        \Expect[\bm{x}_{-n} \sim \D^{n-1}]{ 
         \ProbC[\rup{f^*(\bm{x}_{-n})}]{\atrank{n}{k}}{\lab{n} = \ell} 
        - \ProbC[\rup{f(\bm{x}_{-n})}]{\atrank{n}{k}}{\lab{n} = \ell}
        }\right|.
   \end{align*}

    First, by \cref{eqn:rank-prob-only-label} in the first part of \cref{prop:UA-summary},
    \begin{align*}
       \Expect[\bm{x}_{-n} \sim \D^{n-1}]{ 
       \ProbC[\rup{f^*(\bm{x}_{-n})}]{\atrank{n}{k}}{\lab{n} = \ell}}
       & = 
       \sum_{j=0}^{n-1} \frac{1}{j+1} \cdot 
       \Expect[\bm{x}_{-n} \sim \D^{n-1}]{ 
       \Prob[f^*(\bm{x}_{-n})]{\nlab[-n]{\ell} = j \text{ and } k-(j+1) \leq \nlab[-n]{> \ell} < k}}.
    \end{align*}
    Because the label of $i$ under $f^*$ is drawn by first drawing $x_i \sim \D$, then drawing $\lab{i} \sim f^*(x_i)$, the label distribution of $i$ under $f^*$ is exactly $\bm{p} := \Expect[x \sim \D]{f^*(x)}$ (where we again take expectation of vectors component-wise).
    Writing $P \in \P_{n-1, L}$ for the matrix in which each of the $n-1$ rows is $\bm{p}_i = \bm{p}$, we can therefore write
    \begin{align*}
       \Expect[\bm{x}_{-n} \sim \D^{n-1}]{ 
       \ProbC[\rup{f^*(\bm{x}_{-n})}]{\atrank{n}{k}}{\lab{n} = \ell}}
       & = 
       \sum_{j=0}^{n-1} \frac{1}{j+1} \cdot 
       \Prob[P]{\nlab[-n]{\ell} = j \text{ and } k-(j+1) \leq \nlab[-n]{> \ell} < k}
       \\ & = \ProbC[\rup{P}]{\atrank{n}{k}}{\lab{n} = \ell}.
    \end{align*}
    An identical argument applies for $f$ instead of $f^*$.
    Writing $\bm{q} := \Expect[x \sim \D]{f(x)}$ and $Q \in \P_{n-1, L}$ for the matrix in which each of the $n-1$ rows is $\bm{q}_i = \bm{q}$, we can write
    \begin{align*}
       \Expect[\bm{x}_{-n} \sim \D^{n-1}]{ 
       \ProbC[\rup{f(\bm{x}_{-n})}]{\atrank{n}{k}}{\lab{n} = \ell}}
       & = \ProbC[\rup{Q}]{\atrank{n}{k}}{\lab{n} = \ell}.
    \end{align*}

    Combining both of these calculations, then applying \cref{lem:rank-bound}, we can bound
    \begin{align*}
        & \left|
        \Expect[\bm{x}_{-n} \sim \D^{n-1}]{ 
         \ProbC[\rup{f^*(\bm{x}_{-n})}]{\atrank{n}{k}}{\lab{n} = \ell} 
        - \ProbC[\rup{f(\bm{x}_{-n})}]{\atrank{n}{k}}{\lab{n} = \ell}} 
        \right|
       \\ & \qquad = \left| 
         \ProbC[\rup{P}]{\atrank{n}{k}}{\lab{n} = \ell}
       - \ProbC[\rup{Q}]{\atrank{n}{k}}{\lab{n} = \ell} \right| 
       \\ & \qquad \leq 2 \cdot \sum_{i = 1}^{n-1} \dTV{\bm{p}_i}{\bm{q}_i}
       \\ & \qquad = 2 (n-1) \cdot \dTV{\bm{p}}{\bm{q}}.
       \\ & \qquad = (n-1) \cdot \sum_{\ell} |p_{\ell} - q_{\ell} |
       \\ & \qquad = (n-1) \cdot \sum_{\ell} \left| \Expect[x \sim \D]{f^*(x)_\ell - f(x)_\ell} \right|.
   \end{align*}

    If $f$ is $(\mathcal{C}, \alpha)$-multiaccurate, because $\X \in \mathcal{C}$, we bound
    \begin{align*}
        \sum_{\ell} \left| \Expect[x \sim \D]{ f^*(x)_{\ell} - f(x)_{\ell} } \right|
        & = \sum_{\ell} \left| \Expect[x \sim \D]{ \1_{x \in \X} \cdot \left( f^*(x)_{\ell} - f(x)_{\ell} \right) } \right|
        \; \leq \; L \cdot \alpha.
    \end{align*}
   The identical bound holds for the second part of the theorem, because we assumed $f$ to also be $(\SET{\X}, \alpha)$-multiaccurate.
   
   Finally, we combine all of these bounds, to obtain that
    \begin{align*}
      & \left| \sum_{\ell}
        \left( \Expect[x_n \sim \D]{\1_{x_n \in S} 
        \cdot \left( \Prob[f^*(x_n)]{\lab{n} = \ell} - \Prob[f(x_n)]{\lab{n} = \ell} \right) } \right)
        \cdot \Expect[\bm{x}_{-n} \sim \D^{n-1}]{ 
        \ProbC[\rup{f^*(\bm{x}_{-n})}]{\atrank{n}{k}}{\lab{n} = \ell}
        } \right|
      \\ & \qquad + \left| \sum_{\ell} 
        \Expect[x_n \sim \D]{\1_{x_n \in S} \cdot  
        \Prob[f(x_n)]{\lab{n} = \ell}} \cdot
        \Expect[\bm{x}_{-n} \sim \D^{n-1}]{ 
        \ProbC[\rup{f^*(\bm{x}_{-n})}]{\atrank{n}{k}}{\lab{n} = \ell} 
        - \ProbC[\rup{f(\bm{x}_{-n})}]{\atrank{n}{k}}{\lab{n} = \ell}
        } \right|
    \\ & \leq L \cdot \alpha
        + \sum_{\ell}
        \Expect[x_n \sim \D]{\Prob[f(x_n)]{\lab{n} = \ell}} 
        \cdot  (n-1) \cdot L \cdot \alpha
    \\ & = L \cdot \alpha + (n-1) \cdot L \cdot \alpha
    \; = \; n L \alpha,
   \end{align*}
    completing the proof.
\end{proof}

\cref{prop:multiX-bound} can intuitively be thought of as the following: a predictor which is unbiased on average over a collection of subgroups $\mathcal{C}$ will induce an uncertainty aware ranking which, for those subgroups, has a similar outcome to the (usually inaccessible) uncertainty aware ranking induced by the ground truth label distribution.
The multicalibration guarantee refines this to hold for not only subgroups, but intervals of predictions of the predictor $f$ within that subgroup.
Part (2) of \cref{prop:multiX-bound} requires $f$ be simultaneously $(\mathcal{C}, \alpha, \delta)$-multicalibrated and $(\SET{\X}, \alpha)$-multiaccurate; that is, $f$ is unbiased on average across individuals sampled from $\D$. 
This combination of properties can be achieved by, e.g., the algorithm of \citet{gopalan2022low}.

\subsection{Multicalibration as Interpolating between Group and Individual Fairness}
In contrast to learning accurate individual-level estimates (the Bayes optimal predictor), multiaccuracy/multicalibration can be achieved in time and samples polynomial in the number of sets in $\mathcal{C}$, or, more generally, polynomial in measures of complexity of $\mathcal{C}$ such as its VC-dimension \citep{hebert2018multicalibration,gopalan2022low}.

Notice that if we define $\mathcal{C}_\text{Bayes} = \Set{\SET{x}}{x \in \X}$ to be the set of all singleton groups, 
then $(\mathcal{C}_\text{Bayes}, \alpha)$-multiaccuracy guarantees increasingly accurate predictions for all individuals as $\alpha\to 0$, and $\alpha=0$ recovers the Bayes optimal classifier, i.e., the ground truth $f^*$.
By varying the level of granularity of the collection $\mathcal{C}$, the learned $(\mathcal{C}, \alpha)$-multiaccurate or multicalibrated predictor represents a finer or coarser approximation of the Bayes optimal classifier; thus, \cref{prop:multiX-bound} guarantees that the induced ranking effectively \emph{interpolates} between individual and group-level fair rankings at the granularity defined by $\mathcal{C}$.

Nonetheless, as previously noted, it is usually unreasonable to expect multiaccuracy (or multicalibration) at the level of $(\mathcal{C}_\text{Bayes}, \alpha)$ as $\alpha\to0$.
This is due to information and computational constraints: multicalibration algorithms must use $\Omega(\text{poly}(|\mathcal{C}|, 1/\delta, 1/\alpha))$ samples to learn a multiaccurate/multicalibrated predictor \citep{shabat2020sample}.
In addition to the sample complexity requirements, the class $\mathcal{C}$ must be \emph{agnostic PAC learnable} \citep[Section~3.2]{shalev2014understanding}.
This is a stringent requirement which rarely holds for complex collections such as $\mathcal{C}_\text{Bayes}$.
In practice, we envision \cref{prop:multiX-bound} to be used with sufficiently simple classes $\mathcal{C}$, such as conjunctions of categorical features and intervals of numeric features (e.g., ``women in the age range 45--65'').
Working at this level of granularity not only permits efficient algorithms for obtaining multiaccurate/multicalibrated predictors, but also guarantees that the derived rankings will be unbiased for meaningful protected groups of individuals. 

\section{Ranking Functions and Utility}
\label{sec:utility}
Most online marketplaces utilizing rankings and ranking functions are also concerned with utility or revenue.
In this section, we introduce a natural class of utility models inspired by the literature standard \citep{taylor2008softrank}, and prove that the optimal utility ranking function cannot achieve the stability or group fairness guarantees that UA rankings enjoy.
We then show that a simple ranking function $\rmix$ which is a randomization between the utility-optimal and UA ranking function satisfies an \emph{approximate} notion of stability and fairness while simultaneously retaining a utility guarantee. This may be of use to practitioners more broadly.

\begin{definition}[Utility Model]
Let $w_1 \geq w_2 \geq \cdots \geq w_n \geq 0$ be position weights, and $\tau: \Delta_L \to \R_{\geq 0}$ a function we call the \emph{class utility map}, which determines how predictions are mapped to utilities.
The utility $U(P, r)$ of a prediction matrix $P \in \P_{n, L}$ under a ranking function $r$ is 
\begin{equation*}
    \mathbb{E}_{\sigma \sim r(P)}{\left[ \sum_{k=1}^n w_k \cdot \tau(\bm{p}_{\sigma(k)}) \right]}
    = \sum_{i=1}^n \sum_{k=1}^n \left[ r(P)_{i,k} \cdot w_k \cdot \tau(\bm{p}_i) \right].
\end{equation*}
\end{definition}

A particularly natural and common type of class utility map is the expected utility $\tau(\bm{p}) = \sum_{\ell=1}^L v_\ell \cdot p_\ell$, where $v_L > v_{L-1} > \cdots > v_1 \geq 0$ are the utilities for labels $\ell \in [L]$.
Combined with the weights $w_k = 1/\log_2(1 + k)$, this class captures DCG \citep{jarvelin2002cumulated}, for example.

The ranking function which achieves optimal utility will clearly depend on $\tau$; we denote it by $\ropt$.
It can be simply described as the ranking function which deterministically orders the individuals by decreasing values $\tau(\bm{p}_i)$; recall that $\bm{p}_i$ is the \Kth{i} row of $P \in \P_{n,L}$.
We now show that in general, $\ropt$ is not stable, which demonstrates a necessity to trade off notions of utility and stability.

\begin{proposition}
\label{prop:ropt_not_stable}
    Even for binary labels ($L=2$) and expected utility map $\tau$, the utility-maximizing map $\ropt$ is unstable.
\end{proposition}

\begin{proof}
The example is standard in the literature.
Assume that $v_2 > v_1$. For any $\epsilon \in (0, \frac{1}{2})$, define 
\begin{align*}
    P_\epsilon & = \begin{pmatrix}
        \half + \epsilon & \half - \epsilon \\
        \half - \epsilon & \half + \epsilon 
    \end{pmatrix}
    & P'_\epsilon & = \begin{pmatrix}
        \half - \epsilon & \half + \epsilon \\
        \half + \epsilon & \half - \epsilon 
    \end{pmatrix}.
\end{align*}
Then, $\ropt(P_\epsilon)$ deterministically ranks 2 ahead of 1, while $\ropt(P'_\epsilon)$ deterministically ranks 1 ahead of 2.
As a result, 
$\| \ropt(P_\epsilon) - \ropt(P'_\epsilon) \|_\infty = 1$, while $\| P_\epsilon - P'_\epsilon \|_1 = 8 \epsilon \to 0$ as $\epsilon \to 0$.
This proves instability of $\ropt$.
\end{proof}

Next, we show that for an extremely simple class of instances, namely, when there are two types of individuals with uniform distribution, binary labels, identical uniform ground truth distribution over the two labels for both types, and groups which are just the two singleton types, the utility-maximizing ranking function $\ropt$ cannot approach optimal multigroup fairness, never mind how close to perfectly multiaccurate the predictor gets.

\begin{proposition}
\label{prop:ropt_mcb_lb}
    Let $\X = \SET{1, 2}$ be a domain of two ``types'' of individuals.
    Let $\D$ be the uniform distribution over those two types.
    Let $L = 2$, i.e., we consider binary labels, and the ground truth label distribution is $f^*(x) = (\half, \half)$ for both $x \in \SET{1,2}$, i.e., under the ground truth, both types are equally likely to be good and bad.
    Let $\mathcal{C} = \SET{\SET{1}, \SET{2}}$ be the collection of singleton subgroups.

    For any $\alpha \in (0, \half)$, let $f_{\alpha}$ be the predictor with predictions $f_{\alpha}(1) = (\half-\alpha, \half+\alpha)$ and $f_{\alpha}(2) = (\half+\alpha, \half-\alpha)$.
    (That is, $f_{\alpha}$ slightly overestimates the quality of type 1, and slightly underestimates the quality of type 2.)
    Let $\tau$ be any utility map strictly preferring higher labels, i.e., any utility map with $\tau((\half-\alpha, \half+\alpha)) > \tau((\half+\alpha, \half-\alpha))$.
    Let $\ropt$ be any utility-maximizing ranking function for the utility map $\tau$.
    
    Then, $f_{\alpha}$ is $(\mathcal{C}, \alpha/2)$-multiaccurate, yet for every number $n$ of individuals, the group fairness under $\ropt$ towards the group $S = \SET{1}$ for assignment to the top (most valuable) position in the ranking is the following:

    \begin{align}
         \left| 
                \Expect[\bm{x} \sim \mathcal{D}^n, i \sim \text{Unif}({[n]})]{\1_{x_i \in \SET{1}} \cdot \left( 
                \Prob[\ropteq{f^*(\bm{x})}]{\atrank{i}{1}} - \Prob[\ropteq{f_{\alpha}(\bm{x})}]{\atrank{i}{1}}
                \right) }
        \right|
        & = \frac{1}{n} \cdot \left(\half - 2^{-n}\right).
        \label{eqn:multiaccuracy-lower-bound}
    \end{align}
    In particular, for any fixed $n$, the quantity stays bounded away from 0, even as $\alpha \to 0$. 
\end{proposition}

The intuition for \cref{prop:ropt_mcb_lb} is similar to that for \cref{prop:det-ranking-not-stable}: for the utility-maximizing ranking function, an arbitrarily small (but non-zero) predictive mistake can induce large variations in the resulting ranking distribution, preventing it from preserving group fairness of its predictor.

\begin{proof}
    We first verify that $f_{\alpha}$ is $(\mathcal{C}, \alpha/2)$-multiaccurate.
    For $S = \SET{1}$ or $S = \SET{2}$, we have that
    \begin{align*}
        \| \Expect[x \sim \mathcal{D}]{\1_{x \in S} \cdot (f_{\alpha}(x) - f^*(x))} \|_{\infty}
        & \leq \half \cdot \max (| \half - (\half - \alpha) |, | \half - (\half + \alpha)| )
        = \alpha/2.
    \end{align*}

    The rest of the proof focuses on the group fairness analysis, i.e., proving \cref{eqn:multiaccuracy-lower-bound}.
    We first consider the term $\Prob[\ropteq{f^*(\bm{x})}]{\atrank{i}{1}}$.
    First observe that under the ground truth classifier $f^*$, we have that $f^*(\bm{x}) = \half \cdot \1_{n \times 2}$ for all $\bm{x}$; here $\1_{n \times 2}$ denotes the $n \times 2$ all-ones matrix.
    Under this input matrix, $\ropt$ must have some distribution $\bm{q} = (q_1, \ldots, q_n)$ over which individual is assigned the top rank; crucially for our analysis, because the ranking function \emph{only} observes this matrix $\half \cdot \1_{n \times 2}$, it must use the same distribution for all type vectors $\bm{x}$.
    We thus conclude that $\Prob[\ropteq{f^*(\bm{x})}]{\atrank{i}{1}} = q_i$ for all type vectors $\bm{x}$.

    Next, we consider the term $\Prob[\ropteq{f_{\alpha}(\bm{x})}]{\atrank{i}{1}}$.
    Focus on any type vector $\bm{x} \neq (2, 2, 2, \ldots, 2)$, i.e., a vector that has at least one individual of type 1.
    Because $\tau(f_{\alpha}(1)) > \tau(f_{\alpha}(2))$, and $\ropt$ is utility-maximizing for the utility map $\tau$,
    $\ropt(f_{\alpha}(\bm{x}))$ must rank all individuals of type 1 (of whom there is at least one) ahead of all individuals of type 2.
    From this, we obtain that $\sum_{i=1}^n \1_{x_i \in S} \cdot \Prob[\ropteq{f_{\alpha}(\bm{x})}]{\atrank{i}{1}} = 1$ for all $\bm{x} \neq (2, 2, 2, \ldots, 2)$.
    
    Next, we write out the expectation from \cref{eqn:multiaccuracy-lower-bound}.
    We use that the terms $\1_{x_i \in S} = 0$ for all $i$ when $\bm{x} = (2,2,\ldots, 2)$, which allows us to drop this term from the sum.
    We then use that each $\bm{x}$ under the i.i.d.~uniform type distribution is drawn with probability $2^{-n}$, and substitute our preceding calculations for the probabilities. 
    This gives us the following:
    
    \begin{align*}
        & \left| 
           \Expect[\bm{x} \sim \mathcal{D}^n, i \sim \text{Unif}({[n]})]{\1_{x_i \in S} \cdot \left( 
           \Prob[\ropteq{f^*(\bm{x})}]{\atrank{i}{1}} - \Prob[\ropteq{f(\bm{x})}]{\atrank{i}{1}}
           \right) }
          \right|\\
        &\qquad = \left|
          \sum_{\bm{x}} \sum_{i=1}^n \Pr_{\bm{x} \sim \D^n}[\bm{x}] \cdot \frac{1}{n} \cdot \mathbbm{1}_{x_i \in S}
                \cdot \left( \Prob[\ropteq{f^*(\bm{x})}]{\atrank{i}{1}} - \Prob[\ropteq{f(\bm{x})}]{\atrank{i}{1}} \right)
            \right|\\
        &\qquad = \left|
                2^{-n} \cdot \frac{1}{n} \cdot
                \sum_{\bm{x} \neq (2, 2, \ldots, 2)}
                \left( \sum_{i=1}^n
                \1_{x_i \in S} \cdot q_i - 
                \sum_{i=1}^n \1_{x_i \in S} \cdot \Prob[\ropteq{f(\bm{x})}]{\atrank{i}{1}} \right)
            \right| \\
        &\qquad = 2^{-n} \cdot \frac{1}{n} \cdot \left|
                \sum_{\bm{x} \neq (2, 2, \ldots, 2)}
                \left( \left( \sum_{i=1}^n
                \1_{x_i \in S} \cdot q_i \right) - 1 \right)
            \right| \\
        &\qquad = 2^{-n} \cdot \frac{1}{n} \cdot \left|
                \left( \sum_{i=1}^n q_i \cdot \sum_{\bm{x} \neq (2, 2, \ldots, 2)} 
                \1_{x_i \in S} \right)
                - (2^n-1)
            \right| \\
        &\qquad \stackrel{(\star)}{=} 2^{-n} \cdot \frac{1}{n} \cdot \left|
                \left( \sum_{i=1}^n q_i \cdot 2^{n-1} \right)
                - (2^n-1)
            \right| \\
        &\qquad = 2^{-n} \cdot \frac{1}{n} \cdot \left|
                2^{n-1} - (2^n-1) \right| \\
        &\qquad = \frac{1}{n} \cdot (\half - 2^{-n}).
    \end{align*}
    In the step labeled $(\star)$, we used that there are exactly $2^{n-1}$ vectors with $x_i = 1$; the following step used that the $q_i$, defining a probability distribution, sum to 1.
    This completes the proof.
\end{proof}

\subsection{Utility-Stability Tradeoffs}
In both theory and practice, it is often necessary to trade off utility against other desiderata, such as fairness or stability (see, for example, \citet{singh2019policy, pitoura2022fairness}): if achieving fairness/stability comes at a huge price in utility, it may become economically infeasible to implement fair or stable rankings.
In this section, we introduce and discuss a class $\rmix$ of ranking functions which provide a quantifiable tradeoff between the objectives. 
$\rmix$ linearly interpolates between $\rua$ and $\ropt$ with a trade-off parameter $\phi \in [0,1]$, chosen by the ranking/mechanism designer.
We show that this interpolation naturally leads to $\rmix$ satisfying \emph{approximate} stability and fairness, while providing lower-bound guarantees on the utility.
While the proofs are relatively straightforward, we believe that practitioners may find this class of ranking functions useful in practical applications where the stringent requirement of 1-stability may not be necessary.
The interested reader is referred to \citet{singh2021fairness} for additional discussion on how to choose $\phi$ appropriately\footnote{We note that our approximate fairness guarantee in \cref{prop:multiX-bound-approximate} on multiaccuracy/multicalibration with an additive slack is different from the $\phi$-approximate fairness of \citet{singh2021fairness}, which is a multiplicative notion. Indeed, both hold simultaneously for $\rmix$.}.

We first formally define the notion of \emph{approximate} stability.
\begin{definition}
  Fix $n$ and $L$.
  A ranking function $r: \P_{n,L} \to \dsmatrix$ is \emph{($\gamma$, $\alpha$)-approximately stable} if $|| r(P) - r(P') ||_{\infty} \leq \gamma \cdot ||P-P'||_1 + \alpha$ for all predictions $P, P' \in \P_{n,L}$.
\end{definition}
Notice that $(\gamma, 0)$-approximate stability recovers our stability notion from \cref{def:stable_ranking}.
Approximate stability is a relaxation which allows for additive slack in the dependence on $\| P - P' \|_1$.
An additive slack relaxation is natural and akin to, for example, $(\epsilon, \delta)$-differential privacy (when compared to pure differential privacy).
We show that a simple mixture of UA and the optimal utility ranking satisfies the following approximate stability and utility guarantee.
\begin{proposition}
\label{prop:approx-stable}
    Fix a utility map $\tau$.
    Let $\rmix$ be the ranking function which randomizes between $\rua$ (with probability $\phi$), and $\ropt$ (with probability $1-\phi$).
    Then $\rmix$ is $(\phi, 1-\phi)$-approximately stable.
    Furthermore, for any $P \in \P_{n,L}$, we have that $U(P, \rmix) = \phi \cdot U(P, \rua) + (1-\phi) \cdot U(P, \ropt)$.
\end{proposition}
\begin{proof}
    We first show approximate stability.
    For any $P, P' \in \P_{n,L}$, we have the following.
    \begin{align*}
        \| \rmix(P) - \rmix(P') \|_\infty 
        &= \| \phi \rua(P) + (1-\phi) \ropt(P) - \phi \rua(P') - (1-\phi) \ropt(P') \|_\infty\\
        &\leq \phi \| \rua(P) - \rua(P') \|_\infty + (1-\phi) \| \ropt(P) - \ropt(P') \|_\infty\\
        &\leq \phi \| P - P' \|_1 + (1-\phi).
    \end{align*}
    The last line used the 1-stability of $\rua$ (proved in \cref{prop:stability_of_upr}), as well as the fact that the $\| \cdot \|_\infty$-norm difference of doubly stochastic matrices is at most 1.
    
    The claim about utility is simply linearity of expectations.
\end{proof}

It is straightforward to show that a similar approximate fairness guarantee holds for $\rmix$, again due to its linearity.
\begin{proposition} \label{prop:multiX-bound-approximate}
  Let $\mathcal{D}$ be a distribution over individuals $\mathcal{X}$.
  Let $f^*$ be the ground truth distribution of labels, and $f$ a predictor.
  For any $n$, let $\mathcal{D}^n$ be the distribution obtained from drawing a vector of $n$ i.i.d.~samples from $\mathcal{D}$.
  
  Let $\mathcal{C}$ be a collection of sets with $\mathcal{X} \in \mathcal{C}$, the sets of individuals for which the predictor $f$ will be assumed to be multiaccurate/multcalibrated.
  Let $\alpha \geq 0$ be a parameter for how far from fully accurate/calibrated the predictor is allowed to be.
  \begin{enumerate}
     \item If $f$ is $(\mathcal{C}, \alpha)$-multiaccurate, then the following holds for all sets $S \in \mathcal{C}$ and $k \in [n]$: 
        \begin{align*}
            \left| 
                \Expect[\bm{x} \sim \mathcal{D}^n, i \sim \text{Unif}({[n]})]{\1_{x_i \in S} \cdot \left( 
                \Prob[\rmix(f^*(\bm{x}))]{\atrank{i}{k}} - \Prob[\rmix(f(\bm{x}))]{\atrank{i}{k}}
                \right) }
            \right|
            \leq \phi L n \alpha + 1-\phi .
        \end{align*}
    \item Let some interval width $\delta \in (0,1]$ be given, such that $1/\delta$ is an integer. 
    If $f$ is $(\mathcal{C}, \alpha, \delta)$-multicalibrated and $(\SET{\X}, \alpha)$-multiaccurate, then for every set $S \in \mathcal{C}$, vector $(j_1, j_2, \ldots, j_{L}) \in \SET{0, 1, \ldots, 1/\delta-1}^L$, and $k \in [n]$:
    \begin{align*}
       \Big| 
            \mathbb{E}_{\bm{x} \sim \mathcal{D}^n, i \sim \text{Unif}({[n]})} \Big[
                \1_{x_i \in S} \cdot \1_{f(x_i)_{\ell} \in [j_{\ell} \cdot \delta, (j_{\ell}+1) \cdot \delta) \text{ for all } \ell}  \cdot 
                \left( 
                \Prob[\rup{f^*(\bm{x})}]{\atrank{i}{k}} - \Prob[\rup{f(\bm{x})}]{\atrank{i}{k}}
                \right) \Big]
        \Big|
        \leq \phi L n \alpha + 1-\phi.
    \end{align*}
  \end{enumerate}
\end{proposition}
\begin{proof}
    The proof follows from the following computation due to linearity of $\rmix$:
    \begin{align*}
         \Prob[\rmix(f^*(\bm{x}))]{\atrank{i}{k}} - \Prob[\rmix(f(\bm{x}))]{\atrank{i}{k}}
         = \phi \cdot &\Prob[\rua(f^*(\bm{x}))]{\atrank{i}{k}} + (1-\phi) \cdot \Prob[\ropt(f^*(\bm{x}))]{\atrank{i}{k}} \\
         & - \left( \phi \cdot \Prob[\rua(f(\bm{x}))]{\atrank{i}{k}} + (1-\phi)\cdot \Prob[\ropt(f(\bm{x}))]{\atrank{i}{k}} \right)
    \end{align*}
    Applying this decomposition, then applying linearity of expectation, applying the triangle inequality, and then using \cref{prop:multiX-bound} completes the proof of both parts.
\end{proof}

\section{Experiments on Stability and Utility}
\label{sec:experiments}
We ran experiments on the US census data set \texttt{ACS} curated by \citet{ding2021retiring} and the student dropout task \texttt{Enrollment} introduced by \citet{martins2021early} in the UCI data set Repository.

In our experiments, we demonstrate empirically that the stability guarantees of UA rankings hold when using multiclass predictions.
Furthermore, we find that in practice, the stability guarantees offered by UA ranking may be much better than 1-stability (or the 1/2-stability worst-case lower bound in \cref{prop:stab-lb}), and show that the utility loss suffered by $\rua$ is reasonable.
Although our experiments are relatively simplistic and are not the main focus of our work, they demonstrate that UA rankings have relatively good performance in terms of utility: they outperform not only a uniformly random baseline ranking, but even the Plackett-Luce distribution. At the same time, they also retain the provable fairness and stability properties.

\paragraph{Related Experiments.} Previous work \citep{devic23Fairness, singh2021fairness} also contain experiments demonstrating the utility and utility-fairness tradeoff of UA ranking functions. This past work assumed real-valued predictions (as opposed to the multiclass predictions in our work).
\citet{singh2021fairness} actually deployed a paper recommendation system at a large computer science conference using UA rankings to demonstrate the viability of the method in practice.
Furthermore, their experiments on the \texttt{MovieLens} data set \citep{movielens} demonstrate that UA ranking --- corresponding to a fairness parameter of $\phi=1$ in their work --- can achieve nearly 99\% of the optimal utility given by $\ropt$ in some applications.
\citet{devic23Fairness} show the viability of UA rankings in a matching setting, running experiments on an online dating data set.

\subsection{Stability Against SGD Noise in Neural Network Training}
\label{sec:stability-experiments}

Given our focus on the combination of ranking functions with noisy predictions derived from ML-based classifiers, we first investigate experimentally the stability of UA and utility-maximizing rankings under a natural model of prediction noise. In particular, one of the most common sources of noise in predictions is the randomness in the training procedure (such as SGD). We designed a natural experiment by comparing the behavior of ranking functions under predictors learned with different randomly seeded SGD training runs. Our focus is on understanding if or to what extent the stability of UA and other ranking functions will exceed the worst-case theoretical guarantees in such quasi-realistic settings.

First, we describe the data sets.
In \texttt{ACS}, the prediction target is the binary variable of whether a person is employed or not (after filtering to individuals in the age range 16--90). 
For computational reasons, we restrict our experiments to a subset of the data for California with parameters \texttt{survey year}=`2018', \texttt{horizon}=`1-Year' and \texttt{survey}=`person'.
These parameters are standard when using \texttt{ACS} for testing algorithmic fairness methods, due to the large amount of available data. (See, e.g., the GitHub repository of \citet{ding2021retiring}.)
We are left with 378,817 entries, and use an 80/20 train/test split.
In \texttt{Enrollment}, the target is a multiclass variable for whether an individual is an enrolled, graduated, or dropout student.
After cleaning the data, we are left with 4,424 entries, on which we use an 80/20 train/test split.

Since we want to compare the stability of $\rua$ against that of $\ropt$, we next define simple and natural utilities.
For \texttt{ACS}, we take class 2 to correspond to employment, and class 1 to correspond to unemployment (class 2 $\succ$ class 1). We define $\tau(\bm{p}) = p_2$, i.e., the probability of employment.
For \texttt{Enrollment}, we take class 1 to be that the student has dropped out, class 2 to be enrolled, and class 3 to be graduated (class 3 $\succ$ class 2 $\succ$ class 1), and define $\tau(\bm{p}) = p_1 + 2 \cdot p_2 + 3 \cdot p_3$.

We train 30 simple three-layer MLP neural networks on the \texttt{ACS} data set, which we divide into 15 pairs of networks.
Each pair of networks is initialized with the same (random) weight matrix, then trained separately with SGD.
This introduces noise into the final trained neural network weights, and consequently the predictions.
That is, each pair of networks has similar test accuracy, but will output different probabilities on some (or all) individuals.
On \texttt{ACS}, all networks achieve between 75--80\% train and test accuracy.
We perform the identical procedure for the \texttt{Enrollment} data set, where the networks all achieve between 70--75\% train and test accuracy (due to less data being available).

Let $(f_i, g_i)$, for $i = 1, \ldots, 15$, be the classifiers corresponding to a given pair of networks trained from the same initialization but with different noise due to SGD.
To test the stability of the ranking functions $\rua$ and $\ropt$, for each pair $(f_i,g_i)$, we randomly select 30 individuals from the test set as the data set of individuals $\bm{x}$, and obtain the (probabilistic) predictions $P = f_i(\bm{x})$ and $P' = g_i(\bm{x})$.
We then run $\ropt$ and $\rua$ on these two prediction matrices, logging the deviations of the rankings and the resulting value of $\| P - P' \|_1$.
For each pair of networks $(f_i, g_i)$, we repeat this procedure 10 times with different randomly selected subsets of individuals.
In \cref{tab:stability_employment}, we report the average and standard deviation of this experiment.

\begin{table}[ht]
\centering
\begin{tabular}{@{}lll@{}}
\toprule
Quantity & \texttt{ACS} & \texttt{Enrollment} \\ \midrule
 $\| \rup{P} - \rup{P'} \|_\infty $ & $0.011 \pm 0.002$ & $0.012 \pm 0.002$ \\
 $\| \ropt(P) - \ropt(P')\|_\infty $ & $0.947\pm 0.224$ & $0.680 \pm 0.466$ \\
 $\| P - P' \|_1$ & $0.971\pm 0.582$ & $0.653\pm 0.453$ \\ \bottomrule
\end{tabular}
\caption{Measured stability over 30 neural network training runs (15 pairs of networks) for 10 data sets of $n=30$ individuals each. $\infty$-norm of UA deviation being bounded above by $\| P - P' \|_1$ confirms stability of UA (\cref{prop:stability_of_upr}).
Instability of the ranking $\ropt$ is also demonstrated (\cref{prop:ropt_not_stable}).
\label{tab:stability_employment}}
  \vspace{-0.8em}
\end{table}
The results in \cref{tab:stability_employment} demonstrate that $\|P - P'\|_1$ dominates the value of $\| \rup{P} - \rup{P'} \|_\infty$; this behavior persisted through all of our many training runs.
We conclude that UA rankings are extremely stable in the face of noise introduced during the learning of a predictor (drastically surpassing our 1-stability bound).
Our results also confirm that instability of the optimal ranking function $\ropt$ is not only a theoretical possibility, but prevalent when working with real data.
To see this, notice that the mean value of $\| \ropt(P) - \ropt(P') \|_\infty$ is two orders of magnitude larger than for $\rua$. For the \texttt{Enrollment} data set, it even exceeds the mean value of $\| P - P' \|_1$, implying that $\gamma \leq 1$ is impossible. For the \texttt{ACS} data set, the norms are very comparable, meaning that $\gamma \ll 1$ is impossible; in fact, consistent with the large standard deviations, there are multiple instances illustrating that $\gamma$ must be significantly larger than 1 for both data sets.

\subsection{Utility}
Next, we measure the utility attained by the different ranking functions.
The utility map $\tau$ for both data sets is the same one as in \cref{sec:stability-experiments}.
In addition to the two rankings functions of primary interest, we also consider the following two baselines: (1) $\runif$, the ranking function which places the $n$ individuals in uniformly random order; and (2) $\rpl$, the Plackett-Luce (PL) ranking defined by Luce's axiom \citep{plackett1975analysis, luce2012individual}.

The PL model, similar to UA, defines a distribution over rankings.
At a high level, in each iteration $i$, the item for the \Kth{i} position is sampled based on a softmax mapping of all remaining items' relevance scores.
More precisely, in each iteration $t$, let $M_t$ be the set of individuals not yet placed in the ranking, with $M_1 = [n]$ in the first iteration. Then, in each iteration $t = 1, \ldots, n$, individual $i \in M_t$ is placed in position $t$ with probability 
\[\Prob{\atrank{i}{t}} = \frac{\exp{(\tau(\bm{p}_i))}}{\sum_{j \in M_t} \exp{(\tau(\bm{p}_j))}}.\]

To efficiently compute the PL ranking, we use the (now standard) Gumbel trick from \citet{bruch2020stochastic}.
That is, to sample one ranking from the PL ranking distribution $\rpl(P)$, we sort the individuals in decreasing order of $\tau(\bm{p}_i) + \gamma_i$, where each $\gamma_i \sim \text{Gumbel}(0, 1)$ independently.
We average over 100k samples from the PL ranking distribution to compute $\rpl(P)$.

To measure utility, we use the DCG position weights $w_k = 1/\log_2(k+1)$. 
In order to make the scales of the utilities more meaningful in our comparisons, we normalize all utilities to lie in $[0,1]$.
Thereto, let $\rmin$ be the worst-utility ranking, obtained by ordering the individuals by increasing relevance score (i.e., the individual of lowest utility is deterministically placed first). For a ranking function $r$, we compute the normalized utility score as follows:
\begin{align*}
    \tilde{U}(P, r) = \frac{r(P) - \rmin(P)}{\ropt(P) - \rmin(P)}.
\end{align*}

In \cref{tab:utility}, we report the mean and standard deviation over 30 neural network training runs of the normalized utility $\tilde{U}$ for each of the ranking functions discussed above.
For each neural network and associated prediction function $f$, we randomly sample $n=20, 40,$ and $60$ individuals from the test set.
Then, we construct the prediction matrix $P = (f(x_i))_{i \in [n]}$, and report $\tilde{U}(P, r)$ for each ranking function $r$.
We find that UA ranking outperforms the uniform and PL ranking in each experimental instance.
However, UA ranking is \emph{not} guaranteed to always outperform the uniform ranking.
One can carefully construct instances in which the ``safe bet'' individual provides more utility than an individual who has a low probability of being a ``moonshot'' candidate (further discussed in \citet{singh2021fairness}).
Such an instance crucially depends on the specific choice of $\tau$.

\begin{table}[h!]
\centering
\begin{tabular}{@{}llll@{}}
\toprule
 & $n=20$ & $n=40$ & $n=60$\\ \midrule
 $\rua$   & $\mathbf{0.726} \pm 0.027$ & $\mathbf{0.724}\pm 0.027$ & $\mathbf{0.727}\pm 0.020$ \\
 $\rpl$ & $0.616 \pm 0.038$ & $0.621 \pm 0.028$ & $0.624 \pm 0.023$ \\
 $\runif$ & $0.540 \pm 0.043$ & $0.548 \pm 0.029$ & $0.550 \pm 0.030$ \\ 
 \midrule
 $\rua$    & $\mathbf{0.852} \pm 0.030$ & $\mathbf{0.860 }\pm 0.023$ & $\mathbf{0.857} \pm 0.018$ \\
 $\rpl$ & $0.755 \pm 0.041$ & $0.767 \pm 0.027$ & $0.767 \pm 0.025$ \\
 $\runif$  & $0.552 \pm 0.052$ & $0.561 \pm 0.037$ & $0.562 \pm 0.033$ \\ \bottomrule
\end{tabular}
\caption{Normalized utility achieved by $\rua$, $\runif$, and $\rpl$ for $n=20, 40,$ and $60$ random individuals from the test set of \texttt{ACS} (top 3 rows) and \texttt{Enrollment} (bottom 3 rows). Mean/std taken across 30 neural network training runs. UA outperforms the uniform and PL ranking.}
\label{tab:utility}
\end{table}

\section{Conclusions and Future Work}
\label{sec:conclusions}
Stability of ranking functions is a natural desideratum to prevent large deviations arising in rankings from noise in learned classifiers; combined with individually fair predictions, it results in fair rankings. 
Stability is achieved by the natural Uncertainty Aware Ranking Functions, which also preserve multigroup fairness guarantees of their underlying classifiers.

An interesting direction for future work is to find a more general sufficient condition for ranking functions which allows them to inherit properties of multiaccurate/multicalibrated predictors.
In the proof of \cref{prop:multiX-bound}, we crucially make use of the fact that an individual $i$, when competing against a \emph{sampled} dataset, can be thought of as competing against $n-1$ other individuals sampled from a single conditional distribution. We call this property \emph{individual interpolation}, since it allows for the
\emph{linear} properties of multiaccuracy and multicalibration \citep{noarov2023statisticalscope} to compose with the ranking function.
It would be desirable to characterize which other ranking functions satisfy individual interpolation, and whether the property is necessary for guarantees in the vein of \cref{prop:multiX-bound}.

Another important extension is to consider correlations between the sampled labels of different individuals, and whether analogous individual/group fairness guarantees can still be provided in this case.
In practice, predictors used in rankings in applications such as LinkedIn are trained using highly correlated data due to consumer click-through behavior \citep{diciccio2023detection}.
Investigating ways to train group-fair predictors (in the multiaccurate sense) while only relying on non-i.i.d.~examples, and furthermore applying these predictors in rankings with correlated merit distributions, is an important avenue for future work.

\subsubsection*{Acknowledgements}
SD was supported by the Department of Defense (DOD) through the National Defense Science \& Engineering Graduate (NDSEG) Fellowship Program.
This work was also funded in part by NSF awards \#1916153, \#2333448, \#1956435, \#1943584, \#2344925, and \#2239265, and an Amazon Research Award.

\bibliography{bib-macros/names,bib-macros/conferences,references}
\bibliographystyle{abbrvnat}

\newpage
\appendix
\onecolumn

\section{Proof of Key Lemma for Uncertainty Aware Ranking}
\label{appx:ommited-proofs}
The proof of \cref{lem:rank-bound} uses the following lemma, bounding the total variation distance of sums of random variables in terms of the total variation distances of the individual variables.

\begin{lemma}
\label{lem:coupling-sum}
   Let $X_i \sim \bm{p}_i, Y_i \sim \bm{q}_i$ for $i=1, \ldots, n$ be independent categorical random variables, and $X = \sum_{i=1}^n X_i, Y = \sum_{i=1}^n Y_i$.
   Let $\bm{p}, \bm{q}$ be the respective distributions of $X, Y$.
   Then, $\dTV{\bm{p}}{\bm{q}} \leq \sum_{i=1}^n \dTV{\bm{p}_i}{\bm{q}_i}$.
\end{lemma}

\begin{proof}
    Consider a maximal coupling between each $X_i$ and the corresponding $Y_i$.
    By the Coupling Lemma, we then have that $\Prob{X_i \neq Y_i} = \dTV{\bm{p}_i}{\bm{q}_i}$, and $\dTV{\bm{p}}{\bm{q}} \leq \Prob{X \neq Y}$. Now, by a union bound over all $i$, we obtain that
    \[ 
    \dTV{\bm{p}}{\bm{q}} \leq \Prob{X \neq Y} \leq \sum_i \Prob{X_i \neq Y_i} = \sum_i \dTV{\bm{p}_i}{\bm{q}_i},
    \]
    completing the proof.
\end{proof}

\begin{proof}[Proof of \cref{lem:rank-bound}]
    First, by \cref{eqn:rank-prob-only-label} in the first part of \cref{prop:UA-summary},
    \begin{align*}
       \ProbC{\atrank{i}{k}}{\lab{i} = \ell}
       & = 
       \sum_{j=0}^{n-1} \frac{1}{j+1} \cdot \Prob{\nlab[-i]{\ell} = j \text{ and } k-(j+1) \leq \nlab[-i]{> \ell} < k}.
    \end{align*}

   Let 
   \begin{align*}
     B & = \Set{j}{
           \Prob[P]{\nlab[-i]{\ell} = j \text{ and } k-(j+1) \leq \nlab[-i]{> \ell} < k} \geq 
           \Prob[Q]{\nlab[-i]{\ell} = j \text{ and } k-(j+1) \leq \nlab[-i]{> \ell} < k}};
   \end{align*}
   note that $B$ is not a random variable, but simply determined by the distributions $P, Q$. 

    We substitute the characterization \eqref{eqn:rank-prob-only-label} for both $P$ and $Q$, and use the triangle inequality as well as the fact that $\frac{1}{j+1} \leq 1$, to give us that
  \begin{align}
     & \left| \ProbC[\rup{P}]{\atrank{i}{k}}{\lab{i} = \ell} - \ProbC[\rup{Q}]{\atrank{i}{k}}{\lab{i} = \ell}\right| 
  \nonumber \\ & \qquad \leq 
    \sum_{j=0}^{n-1}  
  \left|  \Prob[P]{\nlab[-i]{\ell} = j \text{ and } k-(j+1) \leq \nlab[-i]{> \ell} < k}
  - \Prob[Q]{\nlab[-i]{\ell} = j \text{ and } k-(j+1) \leq \nlab[-i]{> \ell} < k} \right|
    \nonumber \\ & \qquad = \sum_{j=0, j \in B}^{n-1} 
  \left( \Prob[P]{\nlab[-i]{\ell} = j \text{ and } k-(j+1) \leq \nlab[-i]{> \ell} < k}
  - \Prob[Q]{\nlab[-i]{\ell} = j \text{ and } k-(j+1) \leq \nlab[-i]{> \ell} < k} \right)
  \nonumber \\ & \qquad \qquad + \sum_{j=0, j \notin B}^{n-1} 
  \left( \Prob[Q]{\nlab[-i]{\ell} = j \text{ and } k-(j+1) \leq \nlab[-i]{> \ell} < k}
  - \Prob[P]{\nlab[-i]{\ell} = j \text{ and } k-(j+1) \leq \nlab[-i]{> \ell} < k} \right)
   \nonumber \\ & \qquad = 
  \left( \Prob[P]{\nlab[-i]{\ell} \in B \text{ and } k-(\nlab[-i]{\ell}+1) \leq \nlab[-i]{> \ell} < k}
  - \Prob[Q]{\nlab[-i]{\ell} \in B \text{ and } k-(\nlab[-i]{\ell}+1) \leq \nlab[-i]{> \ell} < k} \right)
  \nonumber \\ & \qquad \qquad + \left( \Prob[Q]{\nlab[-i]{\ell} \notin B \text{ and } k-(\nlab[-i]{\ell}+1) \leq \nlab[-i]{> \ell} < k}
  - \Prob[P]{\nlab[-i]{\ell} \notin B \text{ and } k-(\nlab[-i]{\ell}+1) \leq \nlab[-i]{> \ell} < k} \right)
   \nonumber \\ & \qquad = 
  \left| \Prob[P]{\nlab[-i]{\ell} \in B \text{ and } k-(\nlab[-i]{\ell}+1) \leq \nlab[-i]{> \ell} < k}
  - \Prob[Q]{\nlab[-i]{\ell} \in B \text{ and } k-(\nlab[-i]{\ell}+1) \leq \nlab[-i]{> \ell} < k} \right|
  \nonumber \\ & \qquad \qquad + \left| \Prob[P]{\nlab[-i]{\ell} \notin B \text{ and } k-(\nlab[-i]{\ell}+1) \leq \nlab[-i]{> \ell} < k}
  - \Prob[Q]{\nlab[-i]{\ell} \notin B \text{ and } k-(\nlab[-i]{\ell}+1) \leq \nlab[-i]{> \ell} < k} \right|. \label{eqn:star-bound}
  \end{align}

  Consider the (vector-valued) random variable $(\nlab[-i]{> \ell}, \nlab[-i]{\ell})$, and let $\mu, \nu$ denote its distribution under $P, Q$, respectively.

  Because $[\nlab[-i]{\ell} \in B \text{ and } k-(\nlab[-i]{\ell}+1) \leq \nlab[-i]{> \ell} < k]$ and $[\nlab[-i]{\ell} \notin B \text{ and } k-(\nlab[-i]{\ell}+1) \leq \nlab[-i]{> \ell} < k]$
  are events that can be expressed in terms of this random variable, the definition of total variation distance implies that 
  \begin{align*}
  \left| \Prob[P]{\nlab[-i]{\ell} \in B \text{ and } k-(\nlab[-i]{\ell}+1) \leq \nlab[-i]{> \ell} < k}
  - \Prob[Q]{\nlab[-i]{\ell} \in B \text{ and } k-(\nlab[-i]{\ell}+1) \leq \nlab[-i]{> \ell} < k} \right| & \leq \dTV{\mu}{\nu},
 \\ \left| \Prob[P]{\nlab[-i]{\ell} \notin B \text{ and } k-(\nlab[-i]{\ell}+1) \leq \nlab[-i]{> \ell} < k}
  - \Prob[Q]{\nlab[-i]{\ell} \notin B \text{ and } k-(\nlab[-i]{\ell}+1) \leq \nlab[-i]{> \ell} < k} \right| & \leq \dTV{\mu}{\nu}.
  \end{align*}
  
  To bound $\dTV{\mu}{\nu}$, associate with each individual $i'\neq i$ the 2-dimensional (random) vector $\bm{v}_{i'} = (\1_{\lab{i'} > \ell}, \1_{\lab{i'} = \ell})$. 
  Then, $(\nlab[-i]{> \ell}, \nlab[-i]{\ell}) = \sum_{i'\neq i} \bm{v}_{i'}$.

  For a fixed $i' \neq i$, consider the distribution of $\bm{v}_{i'}$ under $\bm{p}_{i'}$ and $\bm{q}_{i'}$.
  The total variation distance between these distributions is at most $\dTV{\bm{p}_{i'}}{\bm{q}_{i'}}$, because the vectors can differ only when the labels of $i'$ differ.
  By \cref{lem:coupling-sum}, we thus obtain that $\dTV{\mu}{\nu} \leq \sum_{i' \neq i} \dTV{\bm{p}_{i'}}{\bm{q}_{i'}}$.

  Substituting this bound back into \eqref{eqn:star-bound}, we now obtain that
   \begin{align*}
       \left| \ProbC[\rup{P}]{\atrank{i}{k}}{\lab{i} = \ell} - \ProbC[\rup{Q}]{\atrank{i}{k}}{\lab{i} = \ell}\right|
      & \leq 
      2 \cdot \sum_{i' \neq i} \dTV{\bm{p}_{i'}}{\bm{q}_{i'}},
    \end{align*}
   completing the proof.
\end{proof}

\end{document}